\documentclass[10pt,twocolumn,letterpaper]{article}

\usepackage[pagenumbers]{cvpr} 

\definecolor{cvprblue}{rgb}{0.21,0.49,0.74}
\usepackage[pagebackref,breaklinks,colorlinks,allcolors=cvprblue]{hyperref}
\usepackage{bm}
\usepackage{algorithm}
\usepackage{algorithmic}
\usepackage{stfloats}
\usepackage{float}
\usepackage{multicol}

\def\paperID{15605} 
\def\confName{CVPR}
\def\confYear{2025}

\title{Bridging the Gap: Aligning Text-to-Image Diffusion Models with Specific Feedback}

\author{
Xuexiang Niu\textsuperscript{1,\footnote {Equal contribution \label{s1}}} \and
Jinping Tang\textsuperscript{1,\footnotemark[1]} \and
Lei Wang\textsuperscript{1,\footnote {Corresponding author \label{s2}}} \and
Ge Zhu\textsuperscript{1,\footnotemark[2]} \and
\textsuperscript{1}School of Computer and Big Data (School of Cyber Security), Heilongjiang  University \quad
\\
niuxuexiang@s.hlju.edu.cn, \{tangjinping, wanglei, zhuge\}@hlju.edu.cn
}

\begin{document}
\maketitle
\footnotetext[1]{Equal contribution}
\footnotetext[2]{Corresponding author}
\begin{abstract}
Learning from feedback has been shown to enhance the alignment between text prompts and images in text-to-image diffusion models. However, due to the lack of focus in feedback content, especially regarding the object type and quantity, these techniques struggle to accurately match text and images when faced with specified prompts. To address this issue, we propose an efficient fine-turning method with specific reward objectives, including three stages. First, generated images from diffusion model are detected to obtain the object categories and quantities. Meanwhile, the confidence of category and quantity can be derived from the detection results and given prompts. Next, we define a novel matching score, based on above confidence, to measure text-image alignment. It can guide the model for feedback learning in the form of a reward function. Finally, we fine-tune the diffusion model by backpropagation the reward function gradients to generate semantically related images. Different from previous feedbacks that focus more on overall matching, we place more emphasis on the accuracy of entity categories and quantities. Besides, we construct a text-to-image dataset for studying the compositional generation, including 1.7 K pairs of text-image with diverse combinations of entities and quantities. Experimental results on this benchmark show that our model outperforms other SOTA methods in both alignment and fidelity. In addition, our model can also serve as a metric for evaluating text-image alignment in other models. All code and dataset are available at \url{https://github.com/kingniu0329/Visions}.
\end{abstract}

\section{Introduction}

Diffusion models \cite{ref38,ref42,ref36,ref27} have shown excellent performance in the field of text-to-image generation. Despite impressive progress, current models frequently produced images that fail to align well with the specified text prompts, especially for the prompts that contain compositional objects. 

Learning from feedback has demonstrated to be an effective strategy to enhance alignment \cite{ref23}.This strategy typically involves two key steps: (1) Defining an appropriate reward function to measure the alignment between text and image. (2) Based on rewards, fine-tuning the diffusion model through reinforcement learning (do not require differentiable rewards) or backpropagation reward function gradients (require differentiable rewards). In general, feedback contained in reward can be divided into three types: human preferences \cite{ref15,ref16,ref17,ref18}, similarity scores between text and images \cite{ref19,ref20,ref21}, and image quality \cite{ref22,ref23,ref24,ref25}. However, human feedback is costly and suffers from limited scalability, which constrain the training scale and speed of the feedback model. On the other hand, feedbacks based on semantic similarity and image quality are not focused enough on the content, which limits their ability to effectively compose multiple objects \cite{ref7,ref43,ref24}. For instance, current models \cite{ref28,ref45} often face challenges when dealing with specified prompts containing weird or unseen subject, especially regarding the specific object categories and quantities, such as one tiger and two lions on a lotus leaf.
To address the issue of misalignment, we propose an efficient fine-turning text-to-image diffusion model with specific feedback. Unlike previous feedbacks that measured the similarity \cite{ref19,ref21} between text and images ambiguously, we place more emphasis on the accuracy of entity categories and quantities to improve compositional generation. Specifically, we first employ the pre-trained stable diffusion model \cite{ref38} to generate images from text prompts. Then, we detect the categories and locations of objects from generated images via a general detector, and compared them with the tokenized prompts processed by dependency parsing \cite{ref46} to obtain the confidence of category and quantity. Next, we introduce a new matching score, derived from above confidence, to evaluate text-image alignment. This score can be served as a Freward function to guide the model for feedback learning. Finally, we fine-tune the diffusion model by integrating the reward into loss function to produce semantically accurate images. Besides, for the current datasets \cite{ref40}, most generated images is based on descriptions of a single object in different scenarios, which is difficult for generative models to enhance compositional ability. To this end, we construct a text-to-image dataset to explore the compositional generation, including 1,700 pairs of text-image with various combinations of entities and quantities. Our contributions are summarized as follows:

(1) We create a text-to-image dataset for studying the compositional generation. To our best knowledge, this dataset is the first of its kind that contains multiple compositions with different object categories and quantities.

(2) We propose an efficient method with specific feedback for aligning text-to-image diffusion model, which can be fine-turned by a differentiable reward function. This model can also be served as a metric for evaluating text-image alignment in other generation models.

(3) The quantitative and qualitative comparisons demonstrate that our method achieves superior performance over other text-to-image models in both alignment and image quality. Especially, our model shows an average improvement of 11.2\% over SD v1.5 model \cite{ref38}  across the three alignment metrics.
\section{Related Work}
\label{2_relatedwordk}

Diffusion models have achieved significant success in text-to-image generation, such as DALL·E \cite{ref1} and Imagen \cite{ref2}. However, it remains challenging to generate images well-aligned with text prompts. To address the alignment issue, recent researches are broadly categorized into three types: Attention-based, Planning-based and Reward-based methods.

Attention-based methods aim to maintain visual consistency during generation by modifying the attention maps to reduce interference and irrelevant features \cite{ref3,ref4,ref5,ref6,ref7,ref8}. Planning-based methods split a compositional prompt into different objects and generate aligned images conditioned on layouts provided by the user or output of LLM \cite{ref9,ref10,ref11,ref12,ref13,ref14}. The focus of our method is on rewards, so we mainly review such models.

Reward-based methods improve alignment by using feedback from image understanding models. These methods involve two main issues. First, constructing the reward function from appropriate feedback model, such as human preferences \cite{ref15,ref16,ref17,ref18}, similarity scores (e.g., CLIP \cite{ref19}, BLIP \cite{ref20}, BLIP-2 \cite{ref21}), and image quality (e.g., JPEG compressibility \cite{ref22}, aesthetic quality \cite{ref23,ref24}, symmetry \cite{ref25}). Each feedback has its own advantages and limitations. In this work, to enhance alignment between generated images and textual descriptions, particularly in categories and quantities, we propose using object detection for feedback and incorporating its prediction into reward function. Unlike methods conditioned on off-the-shelf predictors (e.g., UG \cite{ref26}, FredDoM \cite{ref27}), our method uses prediction results solely for reward construction and does not require additional training of the detector. The second issue is how to utilize the reward function to guide fine-tuning of the diffusion model. Recent methods can be divided into two groups based on whether they require differentiable rewards for fine-tune. Most methods using non-differentiable rewards fine-tune text-to-image model via reward-weighted likelihood \cite{ref18,ref28} or discarding low-reward images \cite{ref29,ref30,ref31}. By formulating the denoising process as a Markov Decision Process, policy gradient methods can be adopted to fine-tune the text-to-image model for specific rewards \cite{ref22,ref32,ref33,ref34} or modifying the input prompts \cite{ref35}. This kind of approach has the advantage of not requiring differentiable rewards, making it suitable for non-differentiable rewards, but it may result in slower convergence \cite{ref25}. In contrast, methods using differentiable rewards fine-tune diffusion model by reweighting its gradient with reward function gradient \cite{ref36,ref6,ref37,ref15,ref25}. The advantage of this approach lies in its ability to specifically and stably update the generation process by optimizing the reward function. In this work, we construct a differentiable reward function by using a pretrained object detection model, and then fine-tune the model to improve compositional generation towards the accuracy of entity categories and quantities.
\section{Methodology}

\begin{figure*}[htbp]
  \centering
  \includegraphics[width=0.90\linewidth]{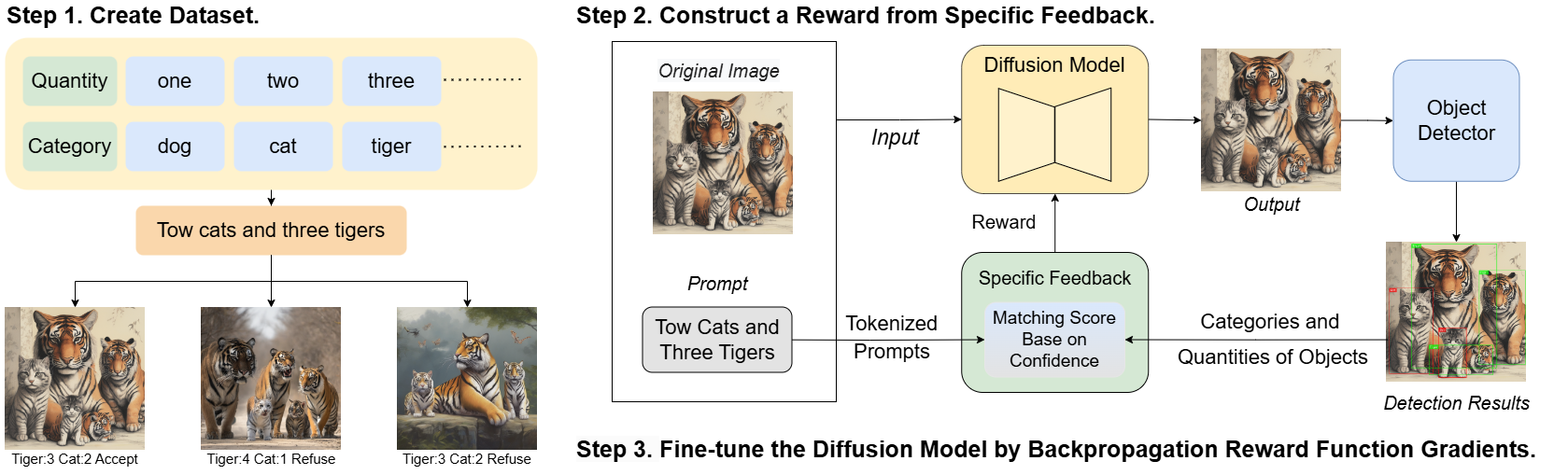} 
  \caption{The steps in our fine-tuning method. (1) We create a text-to-image dataset containing different kinds of compositions. (2) We construct a reward from specific feedback, derived from the confidence in object category and quantity. (3) The diffusion model is fine-tuned by backpropagation reward function gradients to overcome text-image mismatch.}
  \label{fig:ef1}
\end{figure*}

To improve the alignment between the text prompts and generated images, we fine-tune the pre-trained stable diffusion model \cite{ref38} by repeating the following three steps, as shown in Figure \ref{fig:ef1}.

\subsection{Create Dataset \label{zw31}}

To study the compositional problem, our goal is to encourage the model to learn the ability to compose multi-class objects by covering various combinations of categories and quantities. The process mainly involves two steps: (1) text prompts construction, (2) image generation and filtering.

\noindent\textbf{Text prompts construction.} To systematically generate diverse text prompts, we create two types of sets: (1) quantity set and (2) category set. The quantity set consists of words that describe the number of objects, such as {one, two, three, etc}. The category set is made up of words that describe various kinds of objects, such as {bag, fish, tree, etc}. Based on above two sets, we randomly combine the words from them to generate a total of 1,700 text prompts. Notably, when the quantifier is not {one}, the noun will be converted to its plural form, such as “{two} {fishes} and {five} {trees}”. For nouns without an explicitly stated quantity, we default the quantity to one, ensuring completeness and consistency.

\noindent\textbf{Image generation and filtering.} Given a set of text prompts {$\bm{x_1},\bm{x_2},\cdots,\bm{x_k}$}, we first adopt the controllable generation model \cite{ref42} that require additional conditional inputs to produce $n (n>k)$ images {$\bm{z_1},\bm{z_2},\cdots,\bm{z_n}$}. Then the text-image matching scores {$\bm{s_1},\bm{s_2},\cdots,\bm{s_n}$} are calculated by the ImageReward \cite{ref15}, as it has a good correlation with human judgments \cite{ref32}. Next, for each text prompt, we select a set of images with matching scores above a threshold, and then pick out the image that best matches the prompt based on user preferences. This process can effectively filter out images that are inconsistent with the text prompts or of low quality, thereby ensuring the accuracy of the dataset. Finally, we obtain a total of 1,700 pairs of text-image with various combinations of entities and quantities.

\subsection{Construct a reward via specific feedback}

Learning from feedback has emerged as a powerful solution for aligning text-to-image models. However, previous feedbacks \cite{ref13,ref14,ref11} based on semantic similarity are not focused enough on the content, which limits their ability in compositions. To obtain sepecific feedback for alignment, we introduce the detection model to identify the object category and quantity from generated images, and construct a reward function $r(\bm{x},\bm{z})$ from above feedback to fine-tune the diffusion model. The construction of reward function consists of the following three steps. 

\noindent\textbf{First: obtain count and total confidence score of each object from generated image by object detection model.} Given a prompt $\bm{x}$, we first employ the pre-trained stable diffusion model SD v1.5 \cite{ref38} to generate an image $\bm{z}$. Then we input $\bm{z}$ to a object detection model, where YOLOS \cite{ref41} model is considered in this work, as it offers a good balance between efficiency and accuracy.  The outputs of the YOLOS model include the class labels and confidence scores for each bounding box. Bounding boxes with a confidence score below 0.8 are discarded first. Then, Non-Maximum Suppression is applied for boxes with an IoU greater than 0.5 to retain only the box with the highest confidence. 

Finally, we convert the output of YOLOS into a structured key-value format. For example, when the generated image $\bm{x}$ is input into YOLOS model, the outputs are formatted as $\{person : 4; skis : 1\}$. Let ${\bm{z} n}_b$ represents the total number of the remaining detection boxes, ${\bm{z} n}_c$ represents the count of classes labels, $\bm{z}_c^i$ represents the label for each class, and ${\bm{z} n}_b^i$ represents the number of bounding boxes for each class, where the index $i$ ranging from $0$ to ${\bm{z} n}_c$. Then, for this example, the count of detected class is $2$, i.e. $\bm{z} n_c =2$, the detected class labels are $\{person\}$ and $\{skis\}$ respectively, i.e. ${\bm{z}}_c^1 = $“$person”, {\bm{z}}_c^2 = $“$skis”$, and the number of bounding boxes for each class is $4$ and $1$ respectively, i.e. $\bm{z} n_b^1 = 4, \bm{z} n_b^2 = 1$. 

For the $i$-th detected class, $1\leq i\leq {\bm{z} n}_c$, calculate the sum of the confidence score $p_c^i$ of all bounding boxes by $p_c^i = \sum_{k=1}^{{\bm{z} n}_b^i} p_k$, where $p_k$ is the confidence score of the $k$-th bounding box. From the definition of $p_c^i$, it reflecting the overall confidence of the $i$-th class in the image, which also is converted to a structured key-value format. For example, $\{person:3.921;skis:0.903\}$ represents that the total confidence of all bounding boxes for “person” and “skis” is 3.921 and 0.903, respectively, i.e. $p_c^1 = 3.921, p_c^2 = 0.903$.

\noindent\textbf{Second: split textual prompt by nature language model.} In recent years, NLP technologies have made significant progress in understanding and parsing complex texts, particularly in tasks such as part-of-speech tagging and text normalization. To better align with the results processed by the object detection model for matching score, we employ tokenization techniques \cite{ref46} to split the prompt.
First, part-of-speech tagging is applied to each prompt to identify the functional roles of words in the sentence. Then, only quantity words and nouns representing object categories are retained, removing descriptive terms to make the prompt more concise and focused on the core content. To ensure morphological consistency, all nouns are normalized to their singular form (e.g., “dogs” and “dog” are converted to “dog”), reducing matching errors caused by word form variations. After cleaning and simplifying the prompt, we obtain a core prompt containing only quantities and categories. Next, we construct a quantity-category mapping by pairing adjacent quantity words and nouns, forming a clear category-quantity mapping. For nouns without an explicitly stated quantity, we default the quantity to one, ensuring completeness and consistency.

Similar to the first step, to facilitate the calculation of the final matching score, the outputs of the second step are also converted into a structural key-value format. For example, when a textual prompt $\bm{x} =\text{“four person and one skis”}$ is input into the Tokenization model, a structured format $\{person : 4; skis : 1\}$ will be output. Let ${\bm{x} n}_c$ denote the count of classes contained in the prompt, ${\bm{x}}_c^i$ denote the label for each class, and $\bm{x} n_c^i$ denote the number of labels of each class, with the index $i$ ranging from $1$ to ${\bm{x} n}_c$. Then, in this example, the count of classes contained in the prompt is two, i.e. ${\bm{x} n}_c = 2$, the class labels are $\{person\}$ and $\{skis\}$ respectively, i.e. ${\bm{x}}_c^1 = $“$person”, {\bm{x}}_c^2 = $“$skis”$, and the number of labels for each class is $4$ and $1$ respectively, i.e. $\bm{x} n_c^1 = 4, \bm{x} n_c^2 = 1$.

\noindent\textbf{Third: construct reward function by matching score.} With the above processing and notations, we define a novel match score to measure the alignment of text-image, and then integrate the match score into a reward function. The match score primarily consists of two key metrics: the average category confidence and the average quantity confidence. The average category confidence shorted for $\textit{Acc}$, is defined as 
\begin{equation}
	\textit{Acc} = \dfrac{1}{\bm{z} n_c}\sum_{i=1}^{\bm{z} n_c} \dfrac{p_c^i}{\bm{z} n_b^i} * \mathbb{I}\left[\bm{z} _c^i \in{\{\bm{x}_c^j\}_{j=1}^{\bm{x} n_c}} \right], 
\end{equation}
where $\mathbb{I}\left[\bm{z} _c^i \in{\{\bm{x}_c^j\}_{j=1}^{\bm{x} n_c}} \right]$ is the indicator function that equals $1$ when the label $\bm{z}_c^i$ from the generated image $\bm{z}$ lies in the label set $\{\bm{x}_c^j\}_{j=1}^{\bm{x} n_c}$ of the textual prompt $\bm{x}$, and equals $0$ otherwise. The average confidence across all categories is used to measure the accuracy of image category matching. This metric reflects the quality and reliability of generated images across different categories.

The average quantity confidence shorted for $\textit{Aqc}$, is defined as
\begin{equation}
	\label{macro_recall}
	\textit{Aqc} = \dfrac{1}{\bm{z} n_c}\sum_{i=1}^{\bm{z} n_c} \dfrac{1}{\bm{x} n_c} \sum_{j=1}^{\bm{x} n_c}\dfrac{\min\{\bm{z} n_b^i, \bm{x} n_c^j\}}{\max\{\bm{z} n_b^i, \bm{x} n_c^j\}}.
\end{equation}
From the average quantity confidence, it can be found that by calculating the ratio of the minimum to maximum values between the detected count of each category in the generated image and the expected count in the prompt text, and then averaging these ratios across all categories, the confidence in quantity matching of the image can be obtained. This metric primarily evaluates whether the generated image meets the quantitative requirements specified in the text prompt.

Inspired by the F1 score, we define a new matching Score denoted as $CQ\,Score$, in the form of the harmonic mean to balance the contributions of category confidence and quantity confidence to text-image alignment, i.e., 
\begin{equation}
	\label{F1-Score}
	CQ\,Score = \dfrac{2\times\textit{Acc}\times\textit{Aqc}}{\textit{Acc} + \textit{Aqc}}.
\end{equation}

Our $CQ\,Score$ can guide the model for feedback learning in the form of a reward function, i.e., $r(\bm{x},\bm{z}) = CQ\, Score$ for given prompt $\bm{x}$ and the generated image $\bm{z}$. Specifically, $CQ\,Score$ balances the alignment of category and quantity in the generated image, ensuring that the result not only matches the categories in the prompt but also approximates the expected quantity, thus achieve better performance in multi-category and multi-quantity generation tasks.

\subsection{Fine-tune the diffusion model by backpropagation reward function gradients}

Leveraging the reward function defined above, diffusion model can be optimized by backpropagating the reward function gradients to generate semantically related images. There are different ways to apply the reward function to the denoising process, for example, every step of denoising step, randomly selected intermediate step, or the final step. In this work, to directly optimize the final generated image, we apply reward function solely on the last denoising step. The rationale behind this selection is significant: focus on enhancing the final image performance and update the parameters from earlier steps based on the reward signal from the final step to support improvements in the final output.

To unify the goal of maximizing the reward function with the goal of minimizing the loss of diffusion model, we first convert the reward into a reward-driven loss $L_{reward}$, which is defined as 
\begin{equation}
	L_{reward} = \mathbb{E}_{\bm{x} \sim p(\bm{x}), \bm{z} \sim p_{\theta}(\bm{z}|\bm{x})}[\varphi(r(\bm{x},\bm{z}))],
\end{equation}
where $p{(\bm{x})}$ is the distribution of prompt, $p_{\theta}(\bm{z}|\bm{x})$ is the distribution of generated images from a diffusion model $\epsilon_{\theta}$, and $\varphi(\cdot)$ is a function mapping the reward function to a loss, usually chosen as the negative of the reward function. 

To optimize $L_{reward}$, many optimization strategies can be used, for example, reinforce learning and approximate gradient with Monte Corlo Markov Chain. For these two approaches, it is not necessary to require a differentiable reward function, since the gradient of the reward function can be computed by sampling noised images generated in the denoising process. However, multi-step sampling may makes this approach memory inefficient and potentially prone to numerical instability. Note that the reward function constructed in this work is differentiable, direct optimization like SGD can be adopted to backpropagate the reward function gradient.

Fine-tuning solely based on the reward model may lead the model overfit to the reward and discount the ability of the initial diffusion model of generate high quality images. Hence, to address the challenges of rapid overfitting and to enhance stability during fine-tuning, a re-weighting strategy is applied to $L_{reward}$ , along with a regularization using the pre-train loss: $L_{pretrain}$ which is defined as:
\begin{equation}
	L_{pretrain} = \mathbb{E}_{\bm{x} \sim p(\bm{x}), \bm{z}\sim p_{\theta}(\bm{z}|\bm{x})}
		\left[||\epsilon - \epsilon_{\theta}(\bm{z}|\bm{x},t)||^2\right],
\end{equation}
where $t\sim U(0,T)$, $\epsilon\sim \mathcal{N}(0,I)$,  and $\epsilon_{\theta}(\bm{z}|\bm{x},t)$ represents the noise image predicted with the diffusion model $\epsilon_{\theta}$ given the textual prompt $\bm{x}$ and the selected denoising step $t$. $L_{pretrain}$ aims to minimize the difference between the model's predicted noise and the real noise, thereby improving the quality of the generated image and its alignment with the target description.

The overall loss function is defined as: 
\begin{equation}
	L = L_{pretrain} + \lambda * L_{reward},
\end{equation}
where $\lambda$ is a weighting factor which balancing the importance of the reward-driven loss in the total loss. By choosing an appropriate $\lambda$, it is possible to balance the standard diffusion loss and the reward-driven loss, ensuring that the model generates high-quality images while meeting specific quality and consistency requirements. The overall optimization for generating images from given textprompts is summarized in Algorithm1(See Appendix \ref{ap2})

    

\section{Experiments}

\subsection{Experimental Setup}

\noindent\textbf{Implementation Details.} Our model is implemented in PyTorch 2.4.1. All experiments are conducted on a server equipped with four NVIDIA 3090 GPUs, each with 32 GB of memory, and an Intel(R) Xeon(R) Gold 6133 CPU running at 2.50 GHz. We adopt Stable Diffusion v1.5 \cite{ref38}, pre-trained on large image-text datasets \cite{ref48,ref24}, as the foundational generative model and further fine-tune it on the proposed dataset described in Section \ref{zw31} . For the detection model, we choose YOLOS \cite{ref41}, trained on the MS-COCO 2017 dataset \cite{ref40}, as it offers a good balance between efficiency and accuracy. We set the learning rate to 1e-5 and use a cumulative batch size of 2. All training and evaluations are conducted at a resolution of 512 $\times$ 512. For each generation task, images are generated at a resolution of 512 $\times$ 512. The model is fine-tuned using half-precision floating-point numbers and the number of denoising steps is set to 40. 
\begin{figure*}[h!]
 \centering
 \begin{minipage}{0.325\textwidth}
 \centering
 \includegraphics[width=\textwidth]{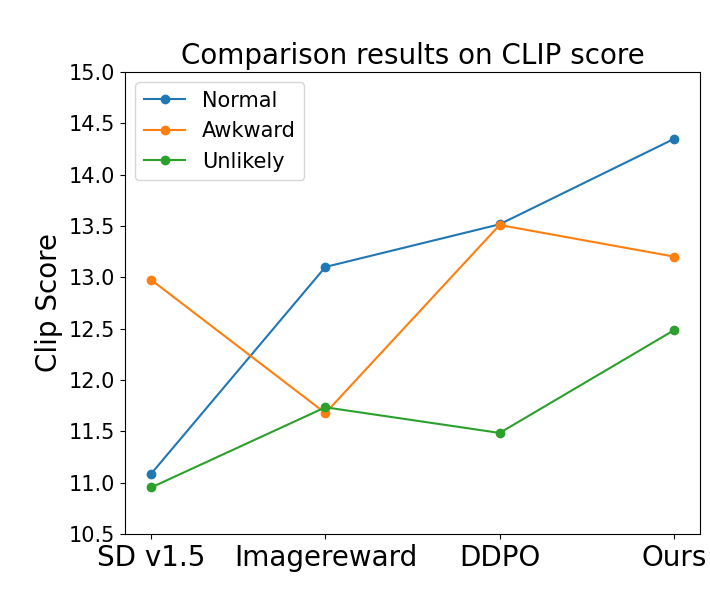}
 \end{minipage}
 \begin{minipage}{0.325\textwidth}
 \centering
 \includegraphics[width=\textwidth]{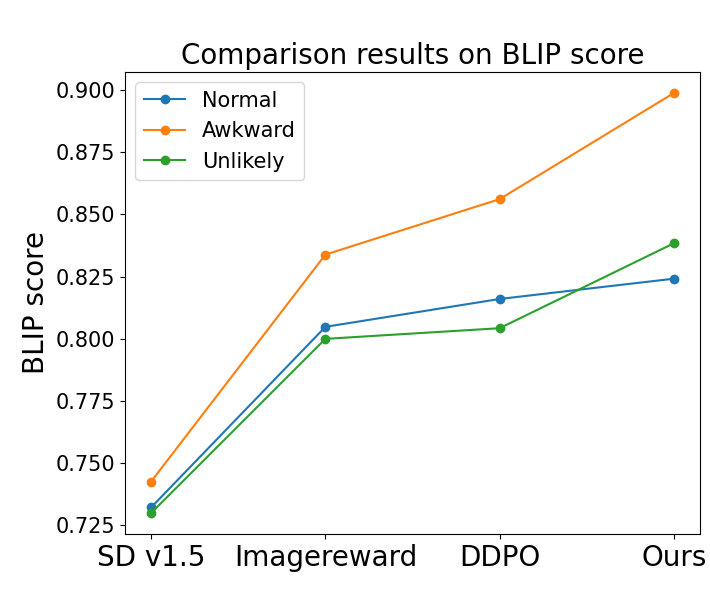}
 \end{minipage}
 \begin{minipage}{0.325\textwidth}
 \centering
 \includegraphics[width=\textwidth]{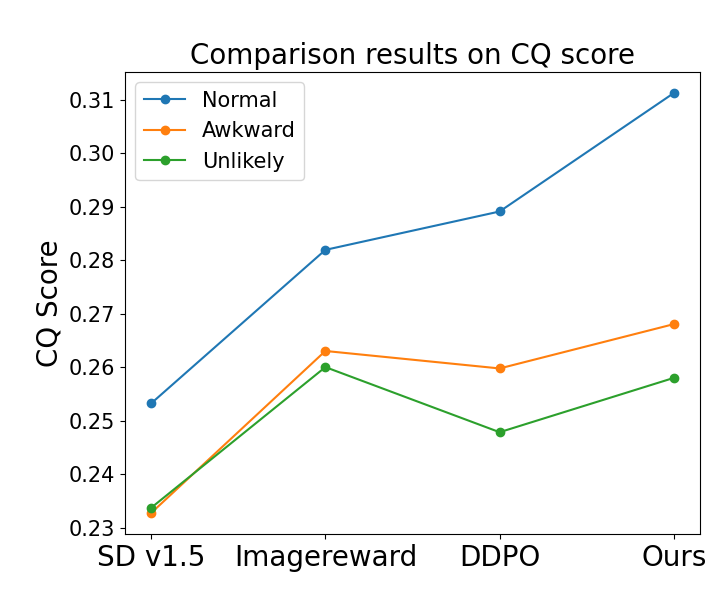}
 \end{minipage}
 
 \centering
 \caption{Comparison results for text-image alignment in three kinds of compositions on three metrics, including \textbf{Normal}, \textbf{Awkward} and \textbf{Unlikely}.}
 \label{fig:f2}
\end{figure*}

\noindent\textbf{Evaluation Metrics.} To comprehensively verify the effectiveness of our model, we evaluated it from two aspects, text-image alignment and quality of generated images. For the alignment, we adopt three metrics to measure the semantic consistency between text prompts and images, including CLIP score\cite{ref19}, BLIP score  \cite{ref20} and the proposed matching score. The higher the above three scores, the better the alignment. For the generation quality, Fr\'{e}chet Inception Distance (FID) \cite{ref47} is employed to assessed the quality of generated images MS-COCO 2017 dataset \cite{ref40}, where a lower FID score represents better image quality.

\subsection{Quantitative Comparison \label{zw42}}

To demonstrate the effectiveness of the proposed feedback strategy, we compare our methods with our baseline model SD v1.5 \cite{ref38} , and two state-of-the-art reward-based models: ImageReward \cite{ref15} and DDPO \cite{ref22}. 

\begin{table}[h!]
 \centering
 \resizebox{0.45\textwidth}{!}{ 
 \begin{tabular}{l|c|c|c|c}
 \hline
 Method & Clip Score ↑ & Blip Score ↑ & CQ Score ↑ & FID ↓\\
 \hline
 SD v1.5 \cite{ref38} & 12.54 & 0.735 & 0.337 & 18.75\\
 ImageReward \cite{ref15} & 13.13 & 0.808 & 0.375 & 18.92 \\
 DDPO \cite{ref22} & 13.11 & 0.793 & 0.371 &18.39 \\
 \hline
 \textbf{Ours} & \textbf{13.42} & \textbf{0.830} & \textbf{0.383} & \textbf{18.38} \\
 \hline
 \end{tabular}
 }
 \caption{Quantitative comparison with baseline model SD v1.5 \cite{ref38}, and other SOTA reward-based models on four evaluation metrics.}
 \label{table:t1}
\end{table}

As shown in Table \ref{table:t1}, our proposed method achieve superior performance over other methods in both alignment and image quality. For the alignment evaluation, the CLIP Score \cite{ref19}, BLIP Score \cite{ref20} and CQ Score of our model improves the original SD v1.5 \cite{ref38} by 7.02\%, 12.93\%, 13.65\%, respectively, which demonstrated our feedback strategy can effectively enhance the semantic consistency between text and images. Furthermore, compared with the ImageReward  \cite{ref15} that constructs reward functions using human preferences, our model improves it by 2.21\%, 2.72\%, 2.13\%, respectively, on the above three metrics. Compared with the DDPO \cite{ref22} that leverages vision-language large models for reward function construction, our method also outperform it, with improvements of 2.36\%, 4.67\%, 3.23\%, respectively. The superior performance of our method can be attributed to our reward perspective, which emphasizes the alignment in category and quantity by introducing specific feedback from object detection. In addition, for the image quality, FID score of our model is also is lower than those of comparison methods, which confirms that our method can generate images that are realistic with the given prompts.

Moreover, we compare with other reward-based models in three kinds of compositions on three alignment metrics, including \textbf{Normal}, \textbf{Awkward} and \textbf{Unlikely}. Among them, \textbf{Normal} represents common composition of objects in daily life, while \textbf{Awkward} indicates the opposite. \textbf{Unlikely} refers to situations that do not exist in reality. As can be seen from Figure \ref{fig:f2}, our method outperforms most of other models on three metrics across three types of compositions, further demonstrating the superiority of our feedback strategy. Only on the \textbf{Awkward}, our method achieved comparable CLIP scores to DDPO \cite{ref22}, as DDOP incorporates aesthetic scores in its rewards for enhancing artistic expression. In summary, our method is superior for multi-object composition tasks. More quantitative analysis on different types of compositions is presented in Appendix \ref{ap1}.

\subsection{Qualitative Comparison}

\begin{figure*}[ht]
 \centering
 \begin{minipage}{0.15\textwidth}
 \centering
 \includegraphics[width=\textwidth,height=1pt]{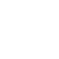}
 \end{minipage}\hfill
 \begin{minipage}{0.187\textwidth}
 \centering
 \textbf{Stable Diffusion\cite{ref38}}
 \end{minipage}\hfill
 \begin{minipage}{0.187\textwidth}
 \centering
 \textbf{ImageReward\cite{ref15}}
 \end{minipage}\hfill
 \begin{minipage}{0.187\textwidth}
 \centering
 \textbf{DDPO\cite{ref22}}
 \end{minipage}\hfill
 \begin{minipage}{0.187\textwidth}
 \centering
 \textbf{Ours}
 \end{minipage}\\[10pt] 


 \begin{minipage}{0.15\textwidth}
 \raggedright
 \textbf{Normal:} \\
 Four cats and two dogs resting on a sunny porch.
 \end{minipage}\hfill
 \begin{minipage}{0.187\textwidth}
 \centering
 \includegraphics[width=\textwidth]{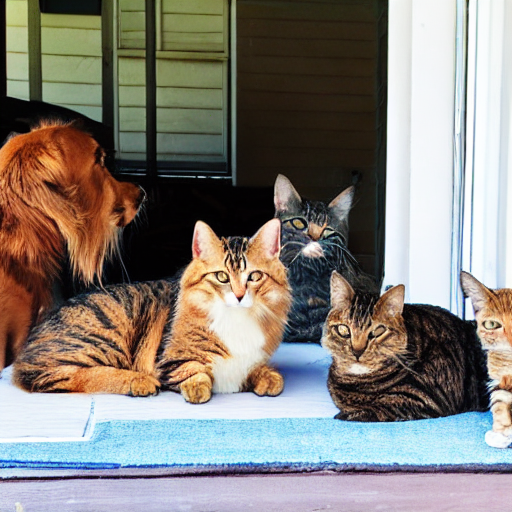}
 \end{minipage}\hfill
 \begin{minipage}{0.187\textwidth}
 \centering
 \includegraphics[width=\textwidth]{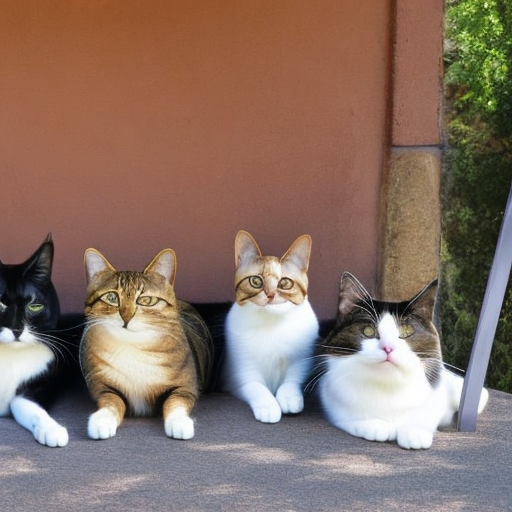}
 \end{minipage}\hfill
 \begin{minipage}{0.187\textwidth}
 \centering
 \includegraphics[width=\textwidth]{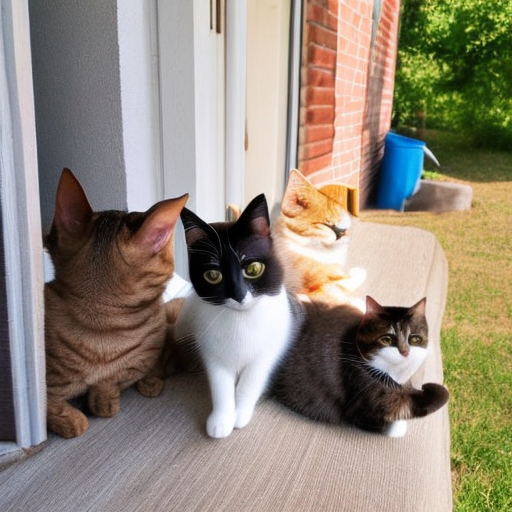}
 \end{minipage}\hfill
 \begin{minipage}{0.187\textwidth}
 \centering
 \includegraphics[width=\textwidth]{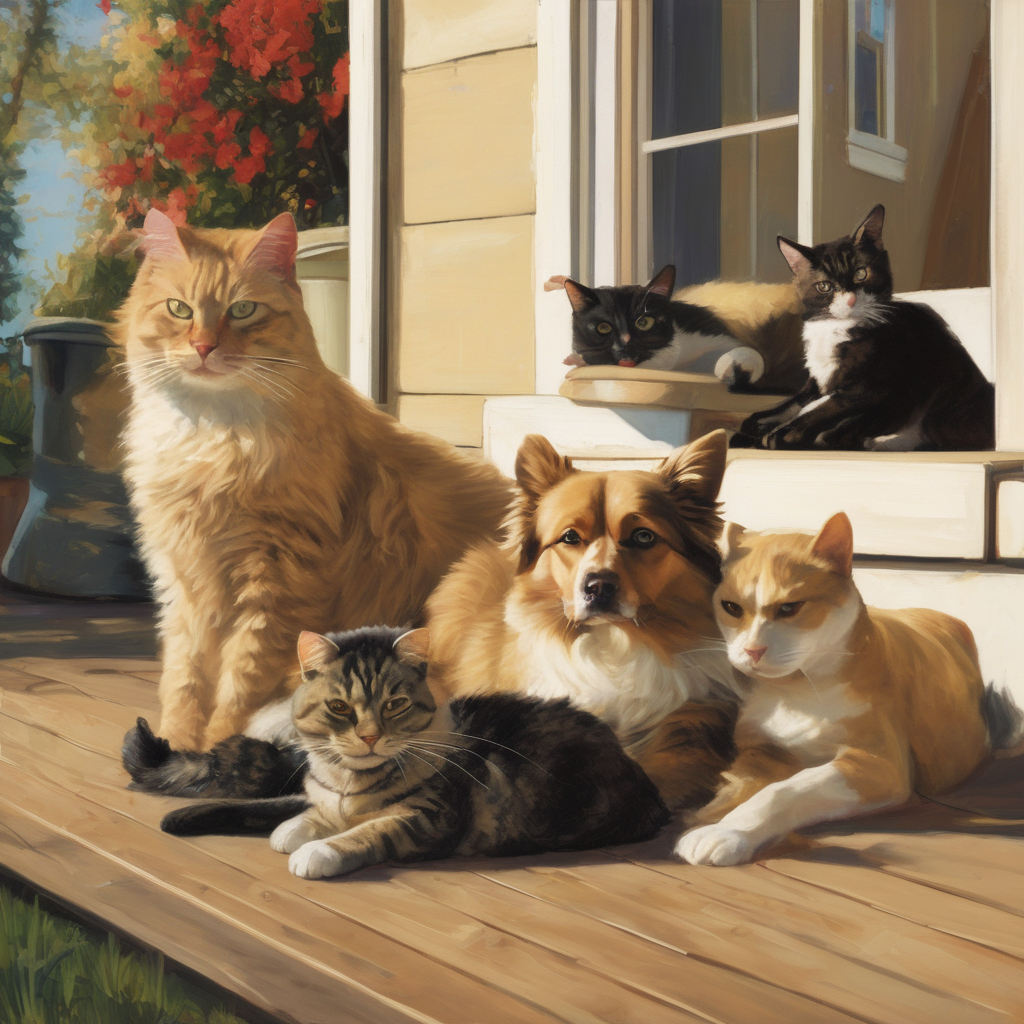}
 \end{minipage}\\[10pt]

 \begin{minipage}{0.15\textwidth}
 \raggedright
 \textbf{Awkward:} \\
 Two dogs and two cats competing in surfing at sea
 \end{minipage}\hfill
 \begin{minipage}{0.187\textwidth}
 \centering
 \includegraphics[width=\textwidth]{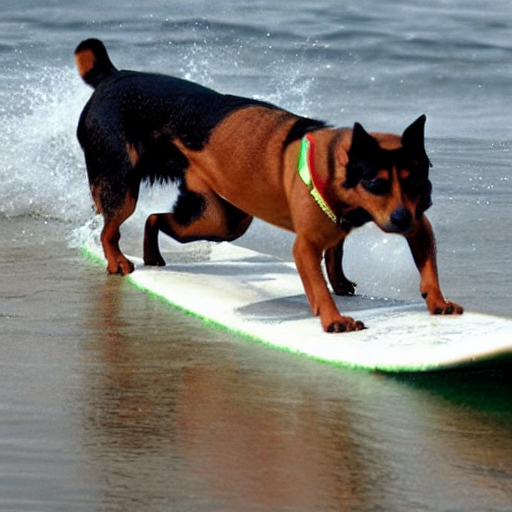}
 \end{minipage}\hfill
 \begin{minipage}{0.187\textwidth}
 \centering
 \includegraphics[width=\textwidth]{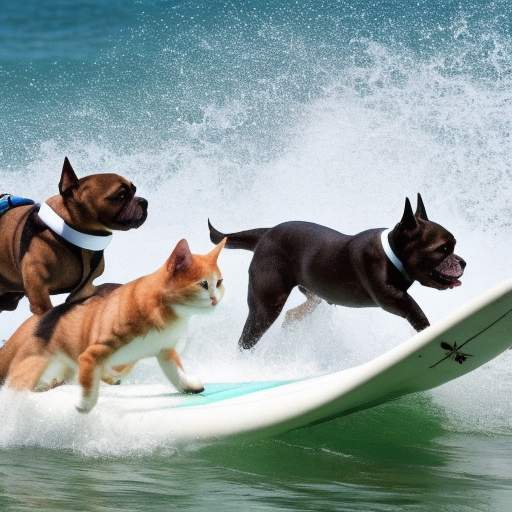}
 \end{minipage}\hfill
 \begin{minipage}{0.187\textwidth}
 \centering
 \includegraphics[width=\textwidth]{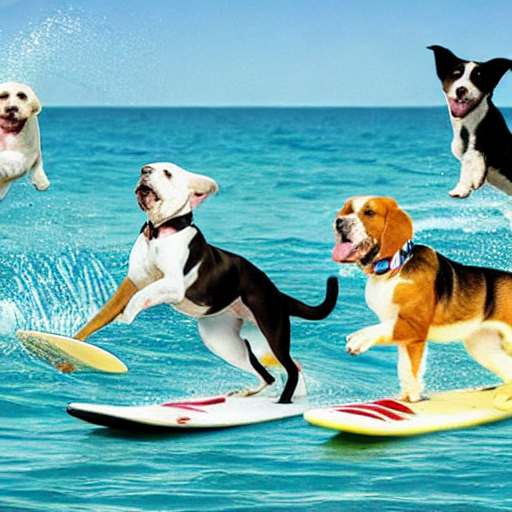}
 \end{minipage}\hfill
 \begin{minipage}{0.187\textwidth}
 \centering
 \includegraphics[width=\textwidth]{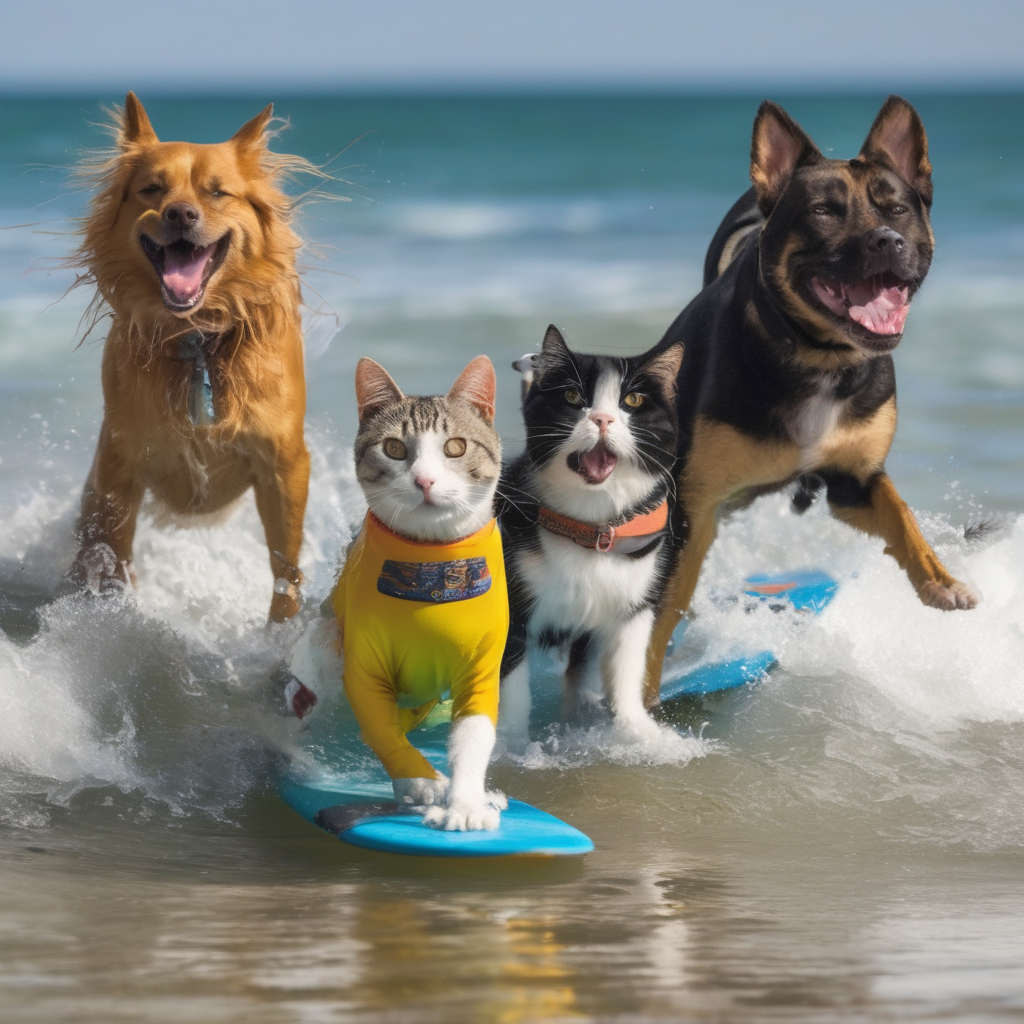}
 \end{minipage}\\[10pt]

 \begin{minipage}{0.15\textwidth}
 \raggedright
 \textbf{Unlikely:} \\
 Three polar bears walking on the moon, wearing spacesuits.
 \end{minipage}\hfill
 \begin{minipage}{0.187\textwidth}
 \centering
 \includegraphics[width=\textwidth]{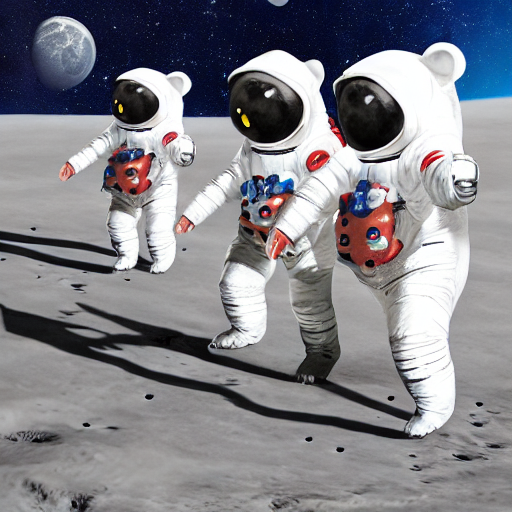}
 \end{minipage}\hfill
 \begin{minipage}{0.187\textwidth}
 \centering
 \includegraphics[width=\textwidth]{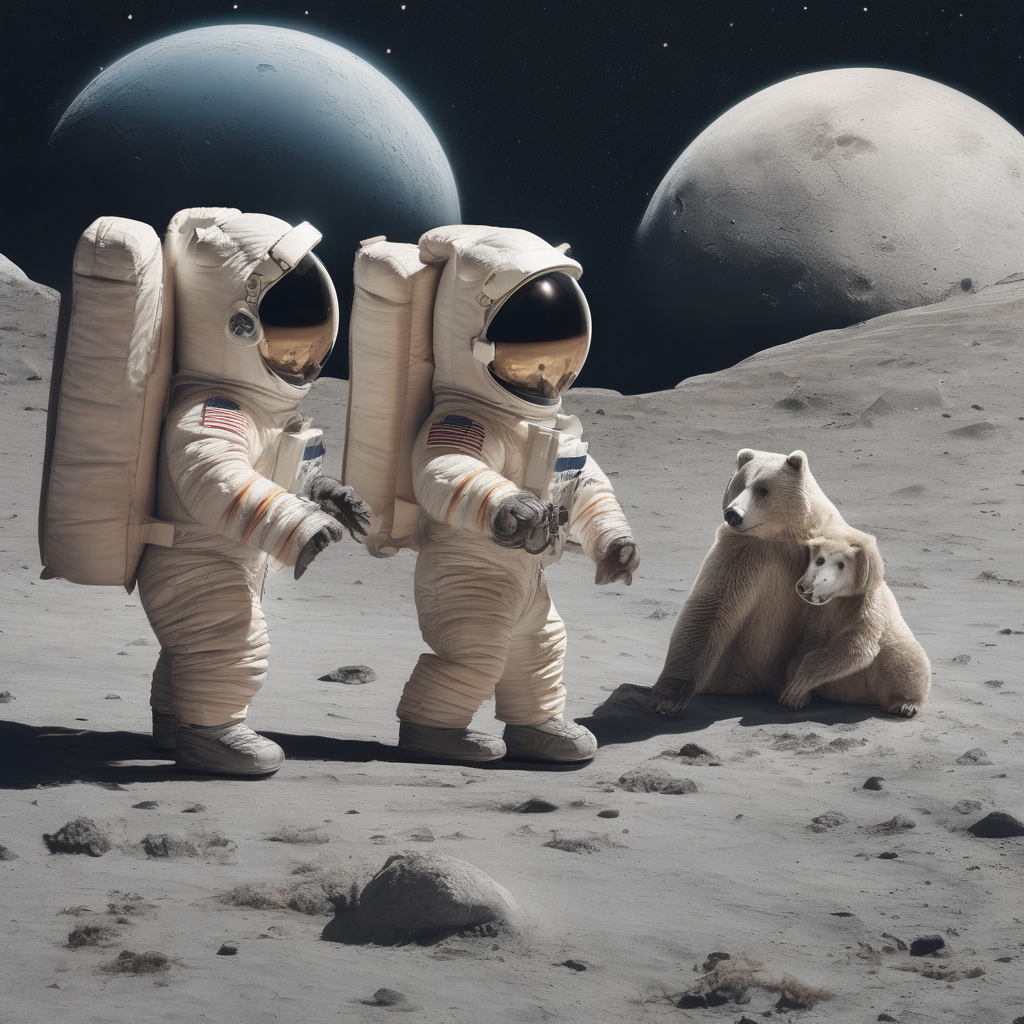}
 \end{minipage}\hfill
 \begin{minipage}{0.187\textwidth}
 \centering
 \includegraphics[width=\textwidth]{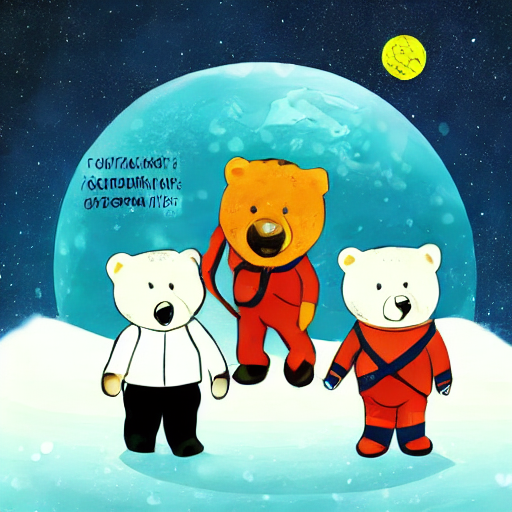}
 \end{minipage}\hfill
 \begin{minipage}{0.187\textwidth}
 \centering
 \includegraphics[width=\textwidth]{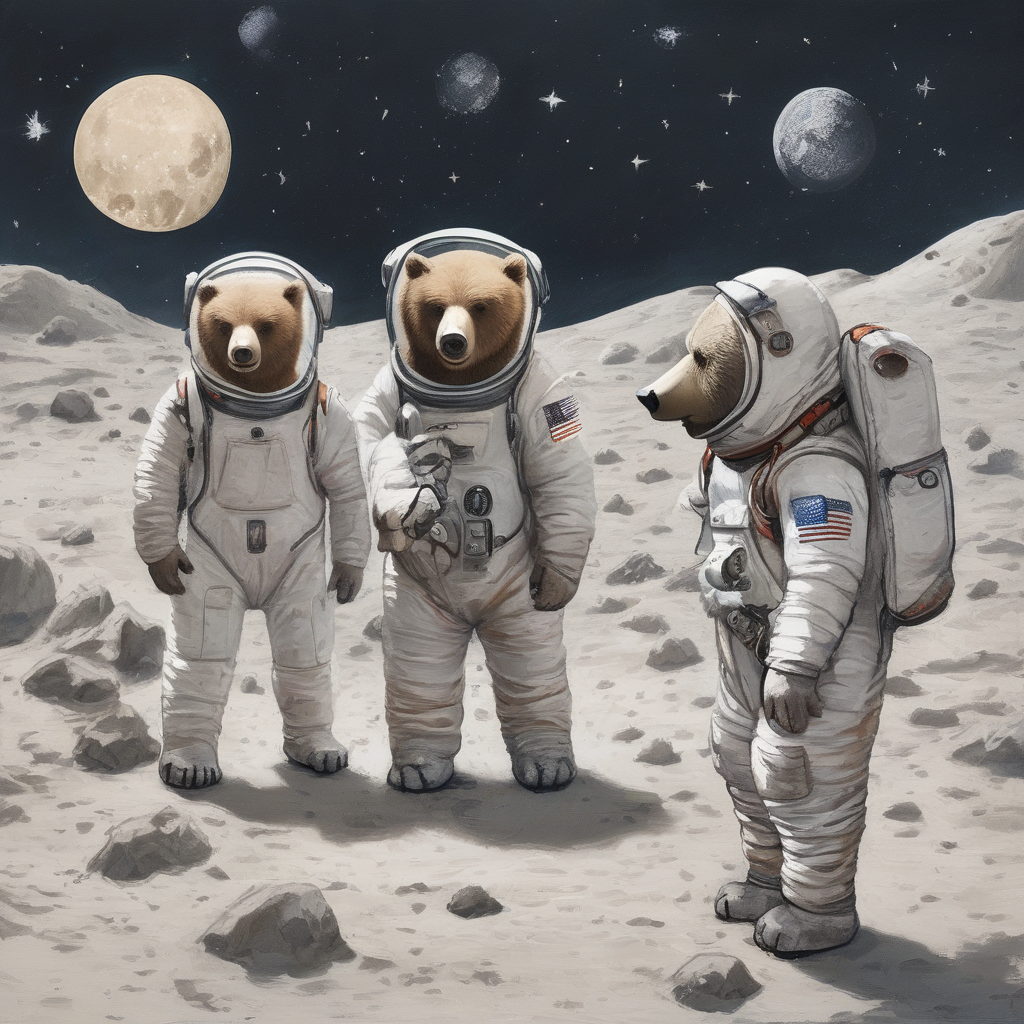}
 \end{minipage}
 
 \caption{Qualitative comparison with three SOTA methods in three kinds of compositions.}
 \label{fig:f3}
\end{figure*}

We present comparison results against the baseline SD v1.5 and other SOTA methods in three kinds of compositions. As can be seen from Figure \ref{fig:f3}, our method achieves a better performance in terms of both image quality and text-image alignment. All the three comparison methods fail to generate images with accurate categories or quantities. For example, for the prompt “Four catsand two dogs resting on a sunny porch”, the baseline SD v1.5 correctly identifies the categories but generates the wrong quantity. This can similarly be seen in the image generated by ImageReward \cite{ref15} with the prompt “Two dogs and two cats competing in surfing at sea”. While DDPO \cite{ref22} fails to generate images with all categories when there are compositions of multi-class objects in the prompt, resulting in a object neglect issue. In contrast, our method is able to synthesize images that more faithfully contain all categories with precise quantities. This indicates that our method can guide the text-to-image model to generate semantically related images by introducing rewards with object attributes.

\section{Conclusion}

In this study, we emphasize the importance of specific feedback for optimizing reward-based text-to-image model. The proposed diffusion model, equipped with an object detector, improves the text-image alignment in two key aspects: category and quantity, both of which are essential for generating semantically relevant images. Besides, our model can also be used as a metric to assess other generation models in matching degree. Moreover, we provided a dataset containing 1,700 pairs of text-image for studying the compositional generation. Experimental results on this dataset demonstrates that our model achieves superior performance over other SOTA methods in both alignment and image quality. We hope our research can inspire the community to explore more precise feedbacks for improving alignment.
{
    \small
    \bibliographystyle{ieeenat_fullname}
    \bibliography{main}

\begin{thebibliography}{46}
\providecommand{\natexlab}[1]{#1}
\providecommand{\url}[1]{\texttt{#1}}
\expandafter\ifx\csname urlstyle\endcsname\relax
  \providecommand{\doi}[1]{doi: #1}\else
  \providecommand{\doi}{doi: \begingroup \urlstyle{rm}\Url}\fi

\bibitem[Agarwal et~al.(2023)Agarwal, Karanam, Joseph, Saxena, Goswami, and Srinivasan]{ref8}
Aishwarya Agarwal, Srikrishna Karanam, K.~J. Joseph, Apoorv Saxena, Koustava Goswami, and Balaji~Vasan Srinivasan.
\newblock {A-STAR:} test-time attention segregation and retention for text-to-image synthesis.
\newblock In \emph{ICCV}, pages 2283--2293, 2023.

\bibitem[Bansal et~al.(2024)Bansal, Chu, Schwarzschild, Sengupta, Goldblum, Geiping, and Goldstein]{ref26}
Arpit Bansal, Hong{-}Min Chu, Avi Schwarzschild, Soumyadip Sengupta, Micah Goldblum, Jonas Geiping, and Tom Goldstein.
\newblock Universal guidance for diffusion models.
\newblock In \emph{ICLR}. OpenReview.net, 2024.

\bibitem[Chefer et~al.(2023)Chefer, Alaluf, Vinker, Wolf, and Cohen{-}Or]{ref5}
Hila Chefer, Yuval Alaluf, Yael Vinker, Lior Wolf, and Daniel Cohen{-}Or.
\newblock Attend-and-excite: Attention-based semantic guidance for text-to-image diffusion models.
\newblock \emph{{ACM} Trans. Graph.}, 42\penalty0 (4):\penalty0 148:1--148:10, 2023.

\bibitem[Chen et~al.(2024)Chen, Laina, and Vedaldi]{ref9}
Minghao Chen, Iro Laina, and Andrea Vedaldi.
\newblock Training-free layout control with cross-attention guidance.
\newblock pages 5331--5341, 2024.

\bibitem[Clark et~al.(2024)Clark, Vicol, Swersky, and Fleet]{ref23}
Kevin Clark, Paul Vicol, Kevin Swersky, and David~J. Fleet.
\newblock Directly fine-tuning diffusion models on differentiable rewards.
\newblock In \emph{ICLR}. OpenReview.net, 2024.

\bibitem[Dong et~al.(2023)Dong, Xiong, Goyal, Zhang, Chow, Pan, Diao, Zhang, Shum, and Zhang]{ref31}
Hanze Dong, Wei Xiong, Deepanshu Goyal, Yihan Zhang, Winnie Chow, Rui Pan, Shizhe Diao, Jipeng Zhang, Kashun Shum, and Tong Zhang.
\newblock {RAFT:} reward ranked finetuning for generative foundation model alignment.
\newblock \emph{Trans. Mach. Learn. Res.}, 2023, 2023.

\bibitem[Fan and Lee(2023)]{ref34}
Ying Fan and Kangwook Lee.
\newblock Optimizing {DDPM} sampling with shortcut fine-tuning.
\newblock pages 9623--9639. {PMLR}, 2023.

\bibitem[Fan et~al.(2023)Fan, Watkins, Du, Liu, Ryu, Boutilier, Abbeel, Ghavamzadeh, Lee, and Lee]{ref32}
Ying Fan, Olivia Watkins, Yuqing Du, Hao Liu, Moonkyung Ryu, Craig Boutilier, Pieter Abbeel, Mohammad Ghavamzadeh, Kangwook Lee, and Kimin Lee.
\newblock {DPOK:} reinforcement learning for fine-tuning text-to-image diffusion models.
\newblock \emph{CoRR}, abs/2305.16381, 2023.

\bibitem[Fang et~al.(2021)Fang, Liao, Wang, Fang, Qi, Wu, Niu, and Liu]{ref41}
Yuxin Fang, Bencheng Liao, Xinggang Wang, Jiemin Fang, Jiyang Qi, Rui Wu, Jianwei Niu, and Wenyu Liu.
\newblock You only look at one sequence: Rethinking transformer in vision through object detection.
\newblock In \emph{NeurIPS}, pages 26183--26197, 2021.

\bibitem[Feng et~al.(2023)Feng, He, Fu, Jampani, Akula, Narayana, Basu, Wang, and Wang]{ref7}
Weixi Feng, Xuehai He, Tsu{-}Jui Fu, Varun Jampani, Arjun~R. Akula, Pradyumna Narayana, Sugato Basu, Xin~Eric Wang, and William~Yang Wang.
\newblock Training-free structured diffusion guidance for compositional text-to-image synthesis.
\newblock In \emph{ICLR}, 2023.

\bibitem[Gokhale et~al.(2022)Gokhale, Palangi, Nushi, Vineet, Horvitz, Kamar, Baral, and Yang]{ref43}
Tejas Gokhale, Hamid Palangi, Besmira Nushi, Vibhav Vineet, Eric Horvitz, Ece Kamar, Chitta Baral, and Yezhou Yang.
\newblock Benchmarking spatial relationships in text-to-image generation.
\newblock \emph{CoRR}, abs/2212.10015, 2022.

\bibitem[Hao et~al.(2023)Hao, Chi, Dong, and Wei]{ref35}
Yaru Hao, Zewen Chi, Li Dong, and Furu Wei.
\newblock Optimizing prompts for text-to-image generation.
\newblock In \emph{NeurIPS}, 2023.

\bibitem[Hessel et~al.(2021)Hessel, Holtzman, Forbes, Bras, and Choi]{ref19}
Jack Hessel, Ari Holtzman, Maxwell Forbes, Ronan~Le Bras, and Yejin Choi.
\newblock Clipscore: {A} reference-free evaluation metric for image captioning.
\newblock pages 7514--7528. Association for Computational Linguistics, 2021.

\bibitem[Honnibal and Montani(2018)]{ref46}
Matthew Honnibal and Ines Montani.
\newblock {spaCy} 2: {Natural} language understanding with {Bloom} embeddings, convolutional neural networks and incremental parsing.
\newblock \emph{To appear}, 2018.

\bibitem[Hu et~al.(2023)Hu, Liu, Kasai, Wang, Ostendorf, Krishna, and Smith]{ref45}
Yushi Hu, Benlin Liu, Jungo Kasai, Yizhong Wang, Mari Ostendorf, Ranjay Krishna, and Noah~A. Smith.
\newblock {TIFA:} accurate and interpretable text-to-image faithfulness evaluation with question answering.
\newblock In \emph{ICCV}, pages 20349--20360, 2023.

\bibitem[Huang et~al.(2023)Huang, Sun, Xie, Li, and Liu]{ref30}
Kaiyi Huang, Kaiyue Sun, Enze Xie, Zhenguo Li, and Xihui Liu.
\newblock T2i-compbench: {A} comprehensive benchmark for open-world compositional text-to-image generation.
\newblock 2023.

\bibitem[Jiang et~al.(2024)Jiang, Song, Wu, Zhang, Shen, Zong, Liu, and Li]{ref6}
Dongzhi Jiang, Guanglu Song, Xiaoshi Wu, Renrui Zhang, Dazhong Shen, Zhuofan Zong, Yu Liu, and Hongsheng Li.
\newblock Comat: Aligning text-to-image diffusion model with image-to-text concept matching.
\newblock \emph{CoRR}, abs/2404.03653, 2024.

\bibitem[Jiang et~al.(2023)Jiang, Fang, Han, Lu, Xu, Liao, Chang, and Liang]{ref36}
Zutao Jiang, Guian Fang, Jianhua Han, Guansong Lu, Hang Xu, Shengcai Liao, Xiaojun Chang, and Xiaodan Liang.
\newblock Realigndiff: Boosting text-to-image diffusion model with coarse-to-fine semantic re-alignment.
\newblock \emph{arXiv preprint arXiv:2305.19599}, 2023.

\bibitem[Kim et~al.(2022)Kim, Kwon, and Ye]{ref37}
Gwanghyun Kim, Taesung Kwon, and Jong~Chul Ye.
\newblock Diffusionclip: Text-guided diffusion models for robust image manipulation.
\newblock In \emph{CVPR}, pages 2416--2425. {IEEE}, 2022.

\bibitem[Kim et~al.(2023)Kim, Lee, Kim, Ha, and Zhu]{ref10}
Yunji Kim, Jiyoung Lee, Jin{-}Hwa Kim, Jung{-}Woo Ha, and Jun{-}Yan Zhu.
\newblock Dense text-to-image generation with attention modulation.
\newblock In \emph{ICCV}, pages 7667--7677, 2023.

\bibitem[Kirstain et~al.(2023)Kirstain, Polyak, Singer, Matiana, Penna, and Levy]{ref17}
Yuval Kirstain, Adam Polyak, Uriel Singer, Shahbuland Matiana, Joe Penna, and Omer Levy.
\newblock Pick-a-pic: An open dataset of user preferences for text-to-image generation.
\newblock In \emph{NeurIPS}, 2023.

\bibitem[Lee et~al.(2020)Lee, Liu, Wu, and Luo]{ref47}
Cheng-Han Lee, Ziwei Liu, Lingyun Wu, and Ping Luo.
\newblock Maskgan: Towards diverse and interactive facial image manipulation.
\newblock In \emph{CVPR}, 2020.

\bibitem[Lee et~al.(2023)Lee, Liu, Ryu, Watkins, Du, Boutilier, Abbeel, Ghavamzadeh, and Gu]{ref28}
Kimin Lee, Hao Liu, Moonkyung Ryu, Olivia Watkins, Yuqing Du, Craig Boutilier, Pieter Abbeel, Mohammad Ghavamzadeh, and Shixiang~Shane Gu.
\newblock Aligning text-to-image models using human feedback.
\newblock \emph{CoRR}, abs/2302.12192, 2023.

\bibitem[Li et~al.(2022)Li, Li, Xiong, and Hoi]{ref20}
Junnan Li, Dongxu Li, Caiming Xiong, and Steven C.~H. Hoi.
\newblock {BLIP:} bootstrapping language-image pre-training for unified vision-language understanding and generation.
\newblock In \emph{International Conference on Machine Learning, {ICML} 2022, 17-23 July 2022, Baltimore, Maryland, {USA}}, pages 12888--12900. {PMLR}, 2022.

\bibitem[Li et~al.(2023{\natexlab{a}})Li, Li, Savarese, and Hoi]{ref21}
Junnan Li, Dongxu Li, Silvio Savarese, and Steven C.~H. Hoi.
\newblock {BLIP-2:} bootstrapping language-image pre-training with frozen image encoders and large language models.
\newblock In \emph{International Conference on Machine Learning, {ICML} 2023, 23-29 July 2023, Honolulu, Hawaii, {USA}}, pages 19730--19742, 2023{\natexlab{a}}.

\bibitem[Li et~al.(2023{\natexlab{b}})Li, Liu, Wu, Mu, Yang, Gao, Li, and Lee]{ref11}
Yuheng Li, Haotian Liu, Qingyang Wu, Fangzhou Mu, Jianwei Yang, Jianfeng Gao, Chunyuan Li, and Yong~Jae Lee.
\newblock {GLIGEN:} open-set grounded text-to-image generation.
\newblock In \emph{CVPR}, pages 22511--22521, 2023{\natexlab{b}}.

\bibitem[Lin et~al.(2014)Lin, Maire, Belongie, Hays, Perona, Ramanan, Doll{\'{a}}r, and Zitnick]{ref40}
Tsung{-}Yi Lin, Michael Maire, Serge~J. Belongie, James Hays, Pietro Perona, Deva Ramanan, Piotr Doll{\'{a}}r, and C.~Lawrence Zitnick.
\newblock Microsoft {COCO:} common objects in context.
\newblock In \emph{ECCV}, pages 740--755. Springer, 2014.

\bibitem[Miao et~al.(2024)Miao, Wang, Wang, Yang, Wang, Qiu, and Liu]{ref22}
Zichen Miao, Jiang Wang, Ze Wang, Zhengyuan Yang, Lijuan Wang, Qiang Qiu, and Zicheng Liu.
\newblock Training diffusion models towards diverse image generation with reinforcement learning.
\newblock In \emph{CVPR}, pages 10844--10853. {IEEE}, 2024.

\bibitem[Ramesh et~al.(2022)Ramesh, Dhariwal, Nichol, Chu, and Chen]{ref1}
Aditya Ramesh, Prafulla Dhariwal, Alex Nichol, Casey Chu, and Mark Chen.
\newblock Hierarchical text-conditional image generation with {CLIP} latents.
\newblock \emph{CoRR}, abs/2204.06125, 2022.

\bibitem[Rassin et~al.(2023)Rassin, Hirsch, Glickman, Ravfogel, Goldberg, and Chechik]{ref4}
Royi Rassin, Eran Hirsch, Daniel Glickman, Shauli Ravfogel, Yoav Goldberg, and Gal Chechik.
\newblock Linguistic binding in diffusion models: Enhancing attribute correspondence through attention map alignment.
\newblock In \emph{NeurIPS}, 2023.

\bibitem[Rombach et~al.(2022)Rombach, Blattmann, Lorenz, Esser, and Ommer]{ref38}
Robin Rombach, Andreas Blattmann, Dominik Lorenz, Patrick Esser, and Bj{\"{o}}rn Ommer.
\newblock High-resolution image synthesis with latent diffusion models.
\newblock In \emph{CVPR}, pages 10674--10685. {IEEE}, 2022.

\bibitem[Saharia et~al.(2022)Saharia, Chan, Saxena, Li, Whang, Denton, Ghasemipour, Lopes, Ayan, Salimans, Ho, Fleet, and Norouzi]{ref2}
Chitwan Saharia, William Chan, Saurabh Saxena, Lala Li, Jay Whang, Emily~L. Denton, Seyed Kamyar~Seyed Ghasemipour, Raphael~Gontijo Lopes, Burcu~Karagol Ayan, Tim Salimans, Jonathan Ho, David~J. Fleet, and Mohammad Norouzi.
\newblock Photorealistic text-to-image diffusion models with deep language understanding.
\newblock In \emph{NeurIPS}, 2022.

\bibitem[Schuhmann et~al.(2021)Schuhmann, Vencu, Beaumont, Kaczmarczyk, Mullis, Katta, Coombes, Jitsev, and Komatsuzaki]{ref48}
Christoph Schuhmann, Richard Vencu, Romain Beaumont, Robert Kaczmarczyk, Clayton Mullis, Aarush Katta, Theo Coombes, Jenia Jitsev, and Aran Komatsuzaki.
\newblock {LAION-400M:} open dataset of clip-filtered 400 million image-text pairs.
\newblock \emph{CoRR}, abs/2111.02114, 2021.

\bibitem[Schuhmann et~al.(2022)Schuhmann, Beaumont, Vencu, Gordon, Wightman, Cherti, Coombes, Katta, Mullis, Wortsman, Schramowski, Kundurthy, Crowson, Schmidt, Kaczmarczyk, and Jitsev]{ref24}
Christoph Schuhmann, Romain Beaumont, Richard Vencu, Cade Gordon, Ross Wightman, Mehdi Cherti, Theo Coombes, Aarush Katta, Clayton Mullis, Mitchell Wortsman, Patrick Schramowski, Srivatsa Kundurthy, Katherine Crowson, Ludwig Schmidt, Robert Kaczmarczyk, and Jenia Jitsev.
\newblock {LAION-5B:} an open large-scale dataset for training next generation image-text models.
\newblock In \emph{NeurIPS}, 2022.

\bibitem[Sun et~al.(2023)Sun, Fu, Hu, Wang, Rassin, Juan, Alon, Herrmann, van Steenkiste, Krishna, and Rashtchian]{ref29}
Jiao Sun, Deqing Fu, Yushi Hu, Su Wang, Royi Rassin, Da{-}Cheng Juan, Dana Alon, Charles Herrmann, Sjoerd van Steenkiste, Ranjay Krishna, and Cyrus Rashtchian.
\newblock Dreamsync: Aligning text-to-image generation with image understanding feedback.
\newblock \emph{CoRR}, abs/2311.17946, 2023.

\bibitem[Wang et~al.(2024)Wang, Sha, Ding, Wang, and Tu]{ref3}
Zirui Wang, Zhizhou Sha, Zheng Ding, Yilin Wang, and Zhuowen Tu.
\newblock Tokencompose: Text-to-image diffusion with token-level supervision.
\newblock In \emph{CVPR}, pages 8553--8564. {IEEE}, 2024.

\bibitem[Wu et~al.(2024{\natexlab{a}})Wu, Lian, Gonzalez, Li, and Darrell]{ref14}
Tsung{-}Han Wu, Long Lian, Joseph~E. Gonzalez, Boyi Li, and Trevor Darrell.
\newblock Self-correcting llm-controlled diffusion models.
\newblock In \emph{CVPR}, pages 6327--6336, 2024{\natexlab{a}}.

\bibitem[Wu et~al.(2023{\natexlab{a}})Wu, Hao, Sun, Chen, Zhu, Zhao, and Li]{ref16}
Xiaoshi Wu, Yiming Hao, Keqiang Sun, Yixiong Chen, Feng Zhu, Rui Zhao, and Hongsheng Li.
\newblock Human preference score v2: {A} solid benchmark for evaluating human preferences of text-to-image synthesis.
\newblock \emph{CoRR}, abs/2306.09341, 2023{\natexlab{a}}.

\bibitem[Wu et~al.(2023{\natexlab{b}})Wu, Sun, Zhu, Zhao, and Li]{ref18}
Xiaoshi Wu, Keqiang Sun, Feng Zhu, Rui Zhao, and Hongsheng Li.
\newblock Human preference score: Better aligning text-to-image models with human preference.
\newblock In \emph{ICCV}, pages 2096--2105. {IEEE}, 2023{\natexlab{b}}.

\bibitem[Wu et~al.(2024{\natexlab{b}})Wu, Hao, Zhang, Sun, Huang, Song, Liu, and Li]{ref25}
Xiaoshi Wu, Yiming Hao, Manyuan Zhang, Keqiang Sun, Zhaoyang Huang, Guanglu Song, Yu Liu, and Hongsheng Li.
\newblock Deep reward supervisions for tuning text-to-image diffusion models.
\newblock \emph{CoRR}, abs/2405.00760, 2024{\natexlab{b}}.

\bibitem[Xie et~al.(2023)Xie, Li, Huang, Liu, Zhang, Zheng, and Shou]{ref12}
Jinheng Xie, Yuexiang Li, Yawen Huang, Haozhe Liu, Wentian Zhang, Yefeng Zheng, and Mike~Zheng Shou.
\newblock Boxdiff: Text-to-image synthesis with training-free box-constrained diffusion.
\newblock In \emph{ICCV}, pages 7418--7427, 2023.

\bibitem[Xu et~al.(2023)Xu, Liu, Wu, Tong, Li, Ding, Tang, and Dong]{ref15}
Jiazheng Xu, Xiao Liu, Yuchen Wu, Yuxuan Tong, Qinkai Li, Ming Ding, Jie Tang, and Yuxiao Dong.
\newblock Imagereward: Learning and evaluating human preferences for text-to-image generation.
\newblock In \emph{NeurIPS}, 2023.

\bibitem[Yang et~al.(2024{\natexlab{a}})Yang, Yu, Meng, Xu, Ermon, and Cui]{ref13}
Ling Yang, Zhaochen Yu, Chenlin Meng, Minkai Xu, Stefano Ermon, and Bin Cui.
\newblock Mastering text-to-image diffusion: Recaptioning, planning, and generating with multimodal llms.
\newblock 2024{\natexlab{a}}.

\bibitem[Yang et~al.(2024{\natexlab{b}})Yang, Chen, and Zhou]{ref33}
Shentao Yang, Tianqi Chen, and Mingyuan Zhou.
\newblock A dense reward view on aligning text-to-image diffusion with preference.
\newblock OpenReview.net, 2024{\natexlab{b}}.

\bibitem[Yu et~al.(2023)Yu, Wang, Zhao, Ghanem, and Zhang]{ref27}
Jiwen Yu, Yinhuai Wang, Chen Zhao, Bernard Ghanem, and Jian Zhang.
\newblock Freedom: Training-free energy-guided conditional diffusion model.
\newblock In \emph{ICCV}, pages 23117--23127. {IEEE}, 2023.

\bibitem[Zhang et~al.(2023)Zhang, Rao, and Agrawala]{ref42}
Lvmin Zhang, Anyi Rao, and Maneesh Agrawala.
\newblock Adding conditional control to text-to-image diffusion models.
\newblock In \emph{ICCV}, pages 3813--3824. {IEEE}, 2023.

\end{thebibliography}
}
\newpage
\onecolumn
\section*{APPENDIX }
\renewcommand{\thesection}{A}
\section{More Experimental Results \label{ap1}}
\label{sec:intro}

\subsection{Analysis on different types of compositions}

\noindent\textbf{Quantitative Comparison.} We design three types of compositions to analyze the effectiveness of different methods on specific combinations of object quantity and category, including (1) Fixed Category \& Incremental Quantity. (2) Random Quantity \& Incremental Category. (3) Incremental Quantity \& Incremental Category. As presented on Table \ref{table:t1}, our method achieves superior performance on alignment metrics for combinations of types 1, 2, and 3. Especially for the type 3, incremental combination of quantity and category, our method has achieved significant improvements compared to the original SD model \cite{ref38}, with an average increase of 9.83\% on three alignment metrics. The improvements across these three combinations demonstrate the focusing ability on semantics of the proposed feedback strategy, which can generate objects with specified category and quantity.
\begin{table}[h!]
 \centering
 \resizebox{0.75\textwidth}{!}{ 
 \begin{tabular}{l|c|c|c}
 \hline
 \multicolumn{4}{c}{Type 1 Fixed Category \& Incremental Quantity.} \\
 \hline
 Method & Clip Score ↑ & Blip Score ↑ & CQ Score ↑ \\
 \hline
 SD v1.5 \cite{ref38} & 23.09 & 0.545 & 0.359 \\
 ImageReward \cite{ref15} & 24.02 & 0.547 & 0.436 \\
 DDPO \cite{ref22} & 24.06 & 0.530 & 0.412 \\
 \hline
 \textbf{Ours} & \textbf{24.18} & \textbf{0.547} & \textbf{0.447} \\
 \hline
 \multicolumn{4}{c}{Type 2 Random Quantity \& Incremental Category.} \\
 \hline
 Method & Clip Score ↑ & Blip Score ↑ & CQ Score ↑ \\
 \hline
 SD v1.5 \cite{ref38}  & 23.94 & 0.367 & 0.117 \\
 ImageReward \cite{ref15} & 24.01 & 0.367 & 0.129 \\
 DDPO \cite{ref22}  & 24.34 & 0.372 & 0.122 \\
 \hline
 \textbf{Ours}   & \textbf{24.73} & \textbf{0.376} & \textbf{0.162} \\
 \hline
 \multicolumn{4}{c}{Type 3 Incremental Quantity \& Incremental Category.} \\
 \hline
 Method & Clip Score ↑ & Blip Score ↑ & CQ Score ↑ \\
 \hline
 SD v1.5 \cite{ref38}  & 23.91 & 0.289 & 0.262 \\
 ImageReward \cite{ref15} & 24.17 & 0.294 & 0.302 \\
 DDPO \cite{ref22}  & 23.72 & 0.297 & 0.277 \\
 \hline
 \textbf{Ours}   & \textbf{25.1} & \textbf{0.297} & \textbf{0.319} \\
 \hline
 \end{tabular}
 }
 \caption{Quantitative comparisons of different types of compositions on three alignment evaluation metrics.
}
 \label{table:t1}
\end{table}

\begin{figure*}[ht]
 \centering
 \begin{minipage}{0.15\textwidth}
 \centering
 \includegraphics[width=\textwidth, height=1pt]{sec/fig/1withe.png}
 \end{minipage}\hfill
 \begin{minipage}{0.187\textwidth}
 \centering
 \textbf{Stable Diffusion\cite{ref38}}
 \end{minipage}\hfill
 \begin{minipage}{0.187\textwidth}
 \centering
 \textbf{ImageReward\cite{ref15}}
 \end{minipage}\hfill
 \begin{minipage}{0.187\textwidth}
 \centering
 \textbf{DDPO\cite{ref22}}
 \end{minipage}\hfill
 \begin{minipage}{0.187\textwidth}
 \centering
 \textbf{Ours}
 \end{minipage}\\[10pt] 

 \begin{minipage}{0.15\textwidth}
 \raggedright
 A sheep on the prairie.
 \end{minipage}\hfill
 \begin{minipage}{0.187\textwidth}
 \centering
 \includegraphics[width=\textwidth]{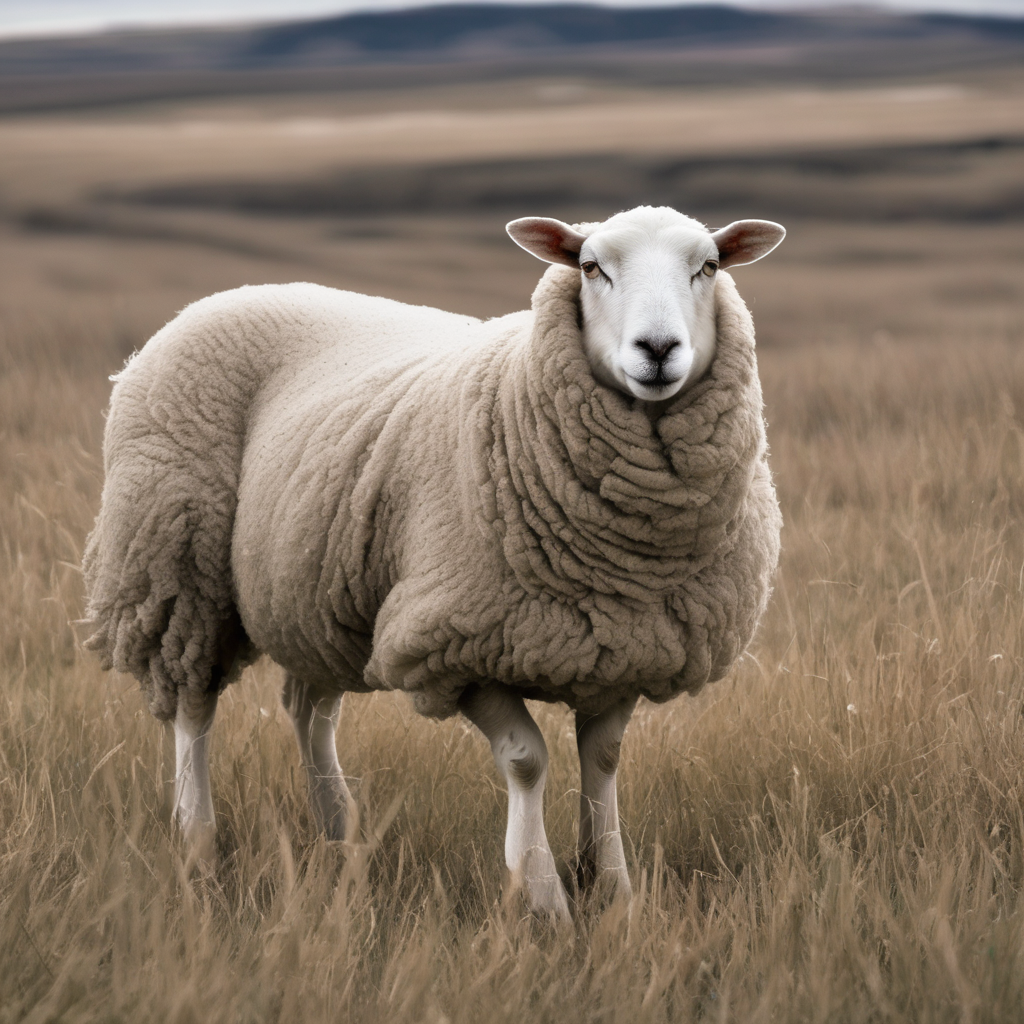}
 \end{minipage}\hfill
 \begin{minipage}{0.187\textwidth}
 \centering
 \includegraphics[width=\textwidth]{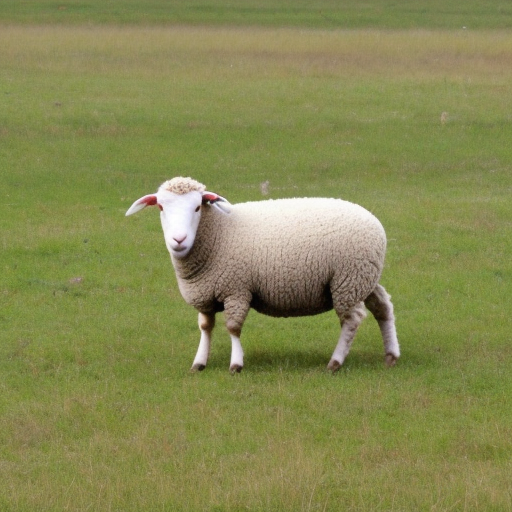}
 \end{minipage}\hfill
 \begin{minipage}{0.187\textwidth}
 \centering
 \includegraphics[width=\textwidth]{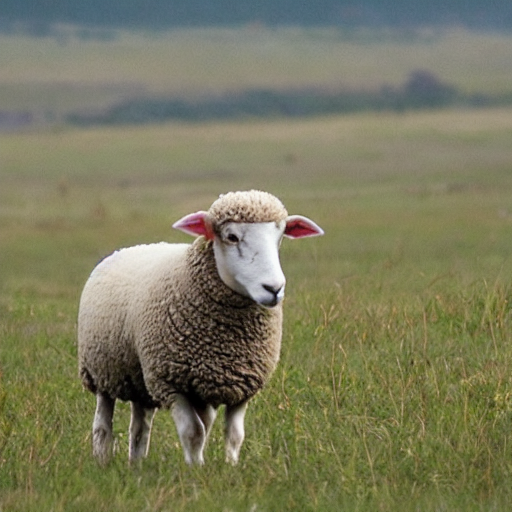}
 \end{minipage}\hfill
 \begin{minipage}{0.187\textwidth}
 \centering
 \includegraphics[width=\textwidth]{sec/fig/cp1/od1.png}
 \end{minipage}\\[10pt]

 \begin{minipage}{0.15\textwidth}
 \raggedright
 Two sheep on the prairie.
 \end{minipage}\hfill
 \begin{minipage}{0.187\textwidth}
 \centering
 \includegraphics[width=\textwidth]{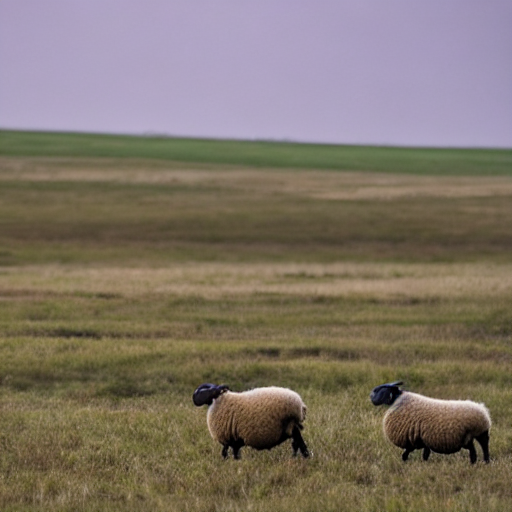}
 \end{minipage}\hfill
 \begin{minipage}{0.187\textwidth}
 \centering
 \includegraphics[width=\textwidth]{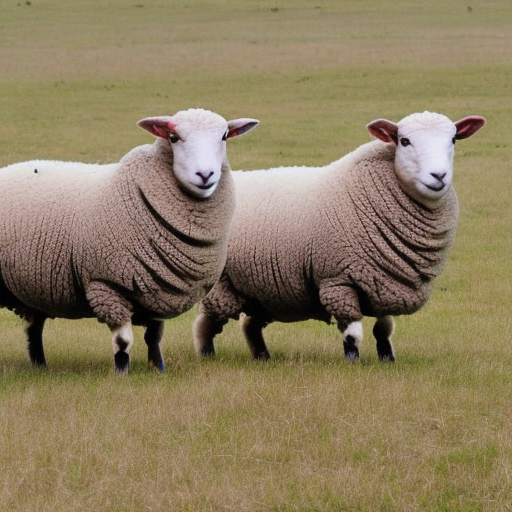}
 \end{minipage}\hfill
 \begin{minipage}{0.187\textwidth}
 \centering
 \includegraphics[width=\textwidth]{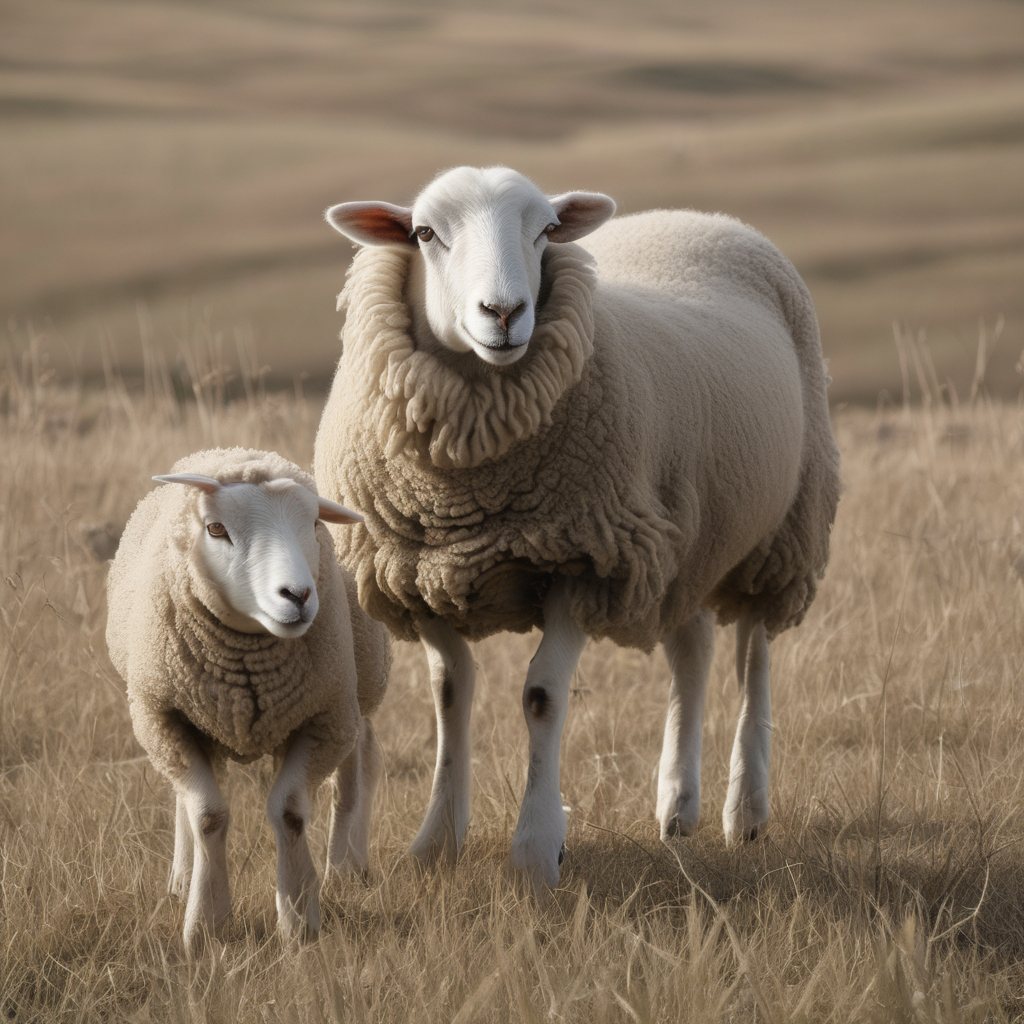}
 \end{minipage}\hfill
 \begin{minipage}{0.187\textwidth}
 \centering
 \includegraphics[width=\textwidth]{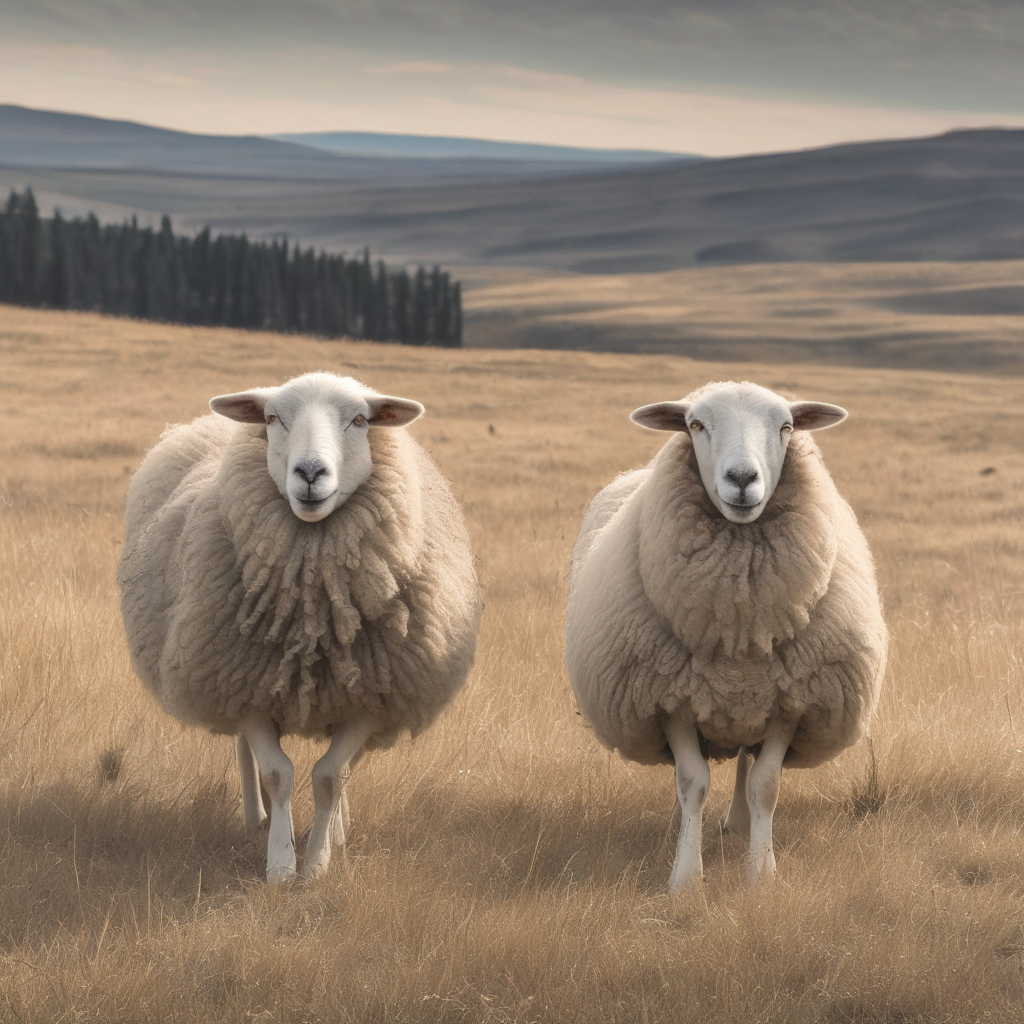}
 \end{minipage}\\[10pt]

 \begin{minipage}{0.15\textwidth}
 \raggedright
 Three sheep on the prairie.
 \end{minipage}\hfill
 \begin{minipage}{0.187\textwidth}
 \centering
 \includegraphics[width=\textwidth]{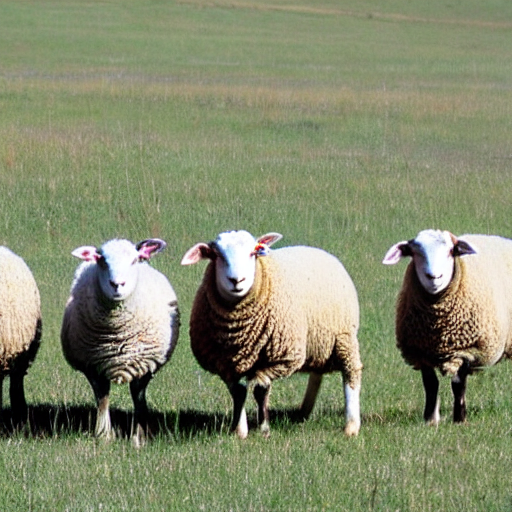}
 \end{minipage}\hfill
 \begin{minipage}{0.187\textwidth}
 \centering
 \includegraphics[width=\textwidth]{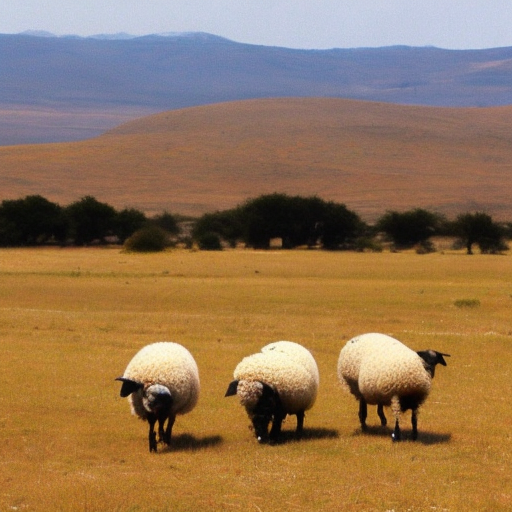}
 \end{minipage}\hfill
 \begin{minipage}{0.187\textwidth}
 \centering
 \includegraphics[width=\textwidth]{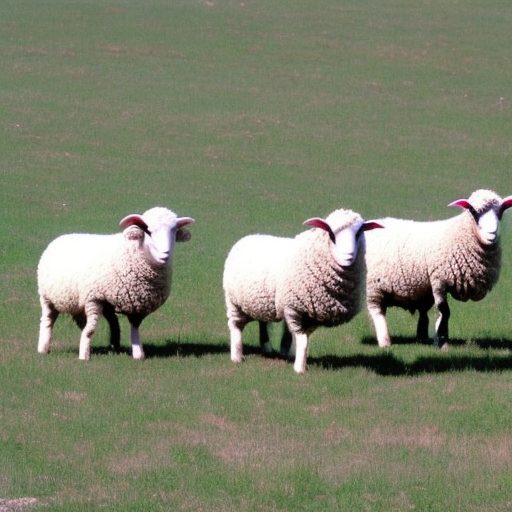}
 \end{minipage}\hfill
 \begin{minipage}{0.187\textwidth}
 \centering
 \includegraphics[width=\textwidth]{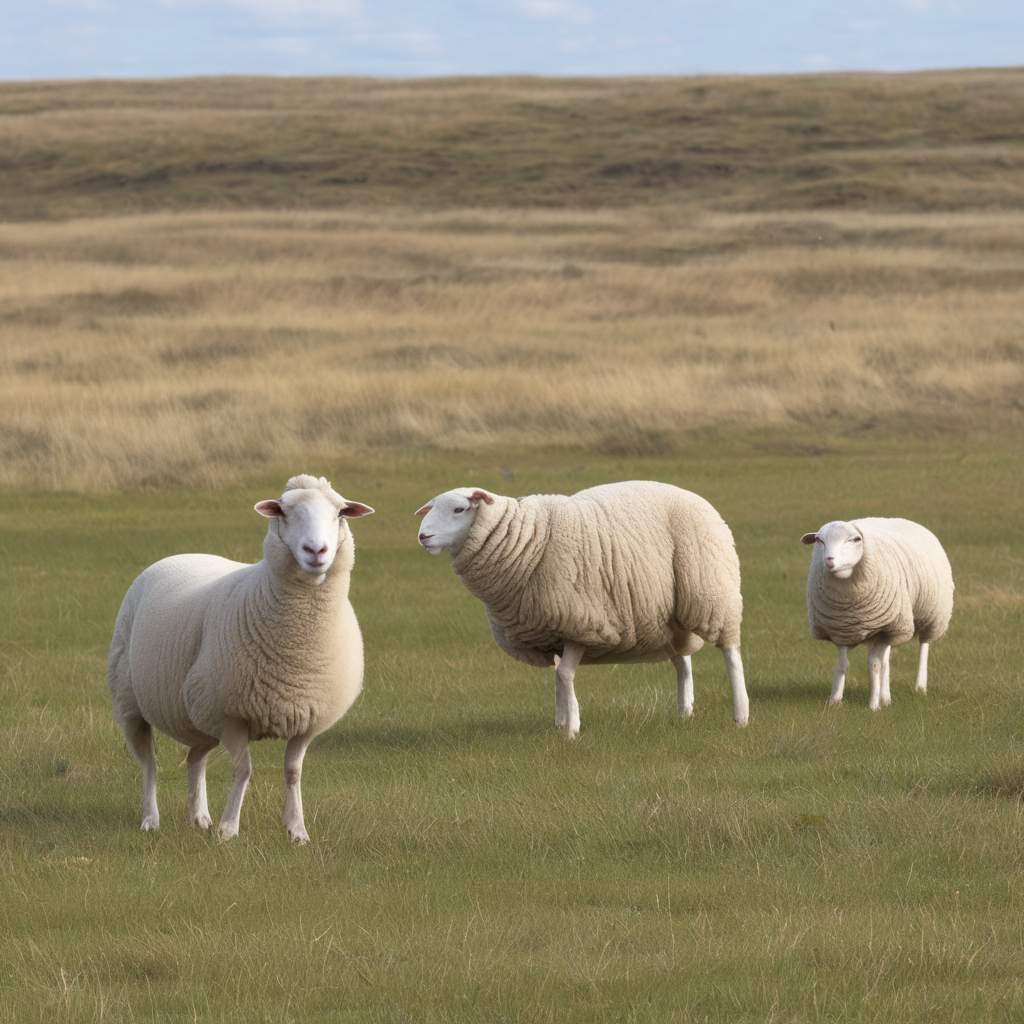}
 \end{minipage}\\[10pt]

 \begin{minipage}{0.15\textwidth}
 \raggedright
 Four sheep on the prairie.
 \end{minipage}\hfill
 \begin{minipage}{0.187\textwidth}
 \centering
 \includegraphics[width=\textwidth]{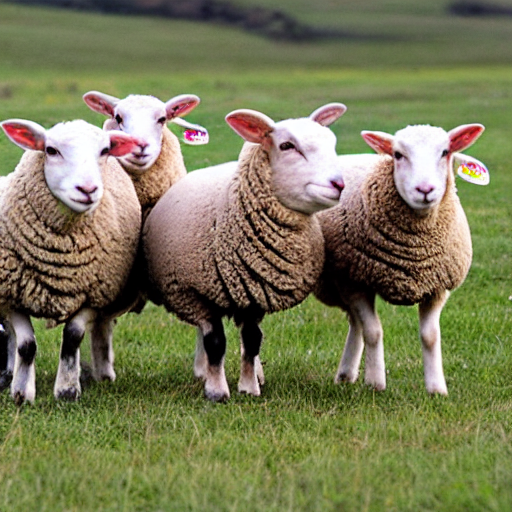}
 \end{minipage}\hfill
 \begin{minipage}{0.187\textwidth}
 \centering
 \includegraphics[width=\textwidth]{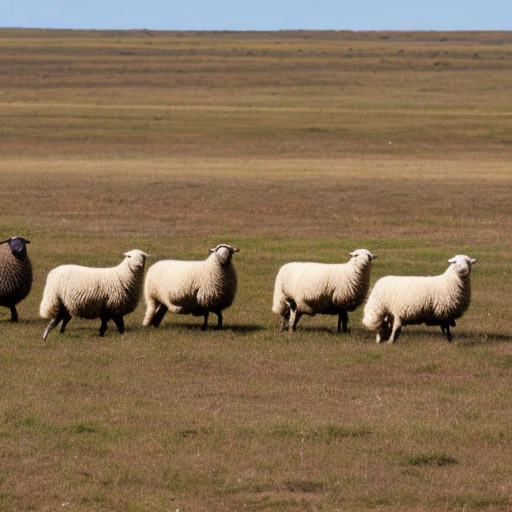}
 \end{minipage}\hfill
 \begin{minipage}{0.187\textwidth}
 \centering
 \includegraphics[width=\textwidth]{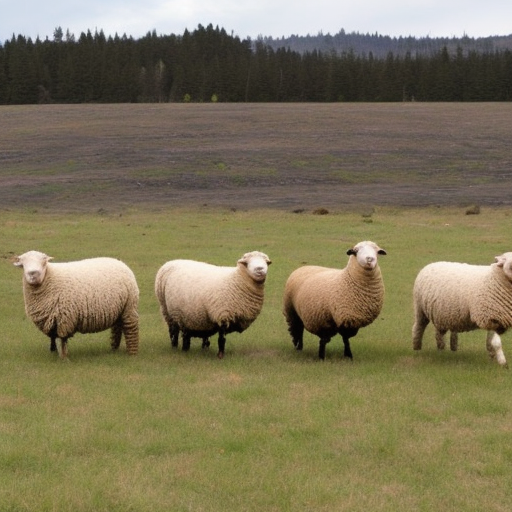}
 \end{minipage}\hfill
 \begin{minipage}{0.187\textwidth}
 \centering
 \includegraphics[width=\textwidth]{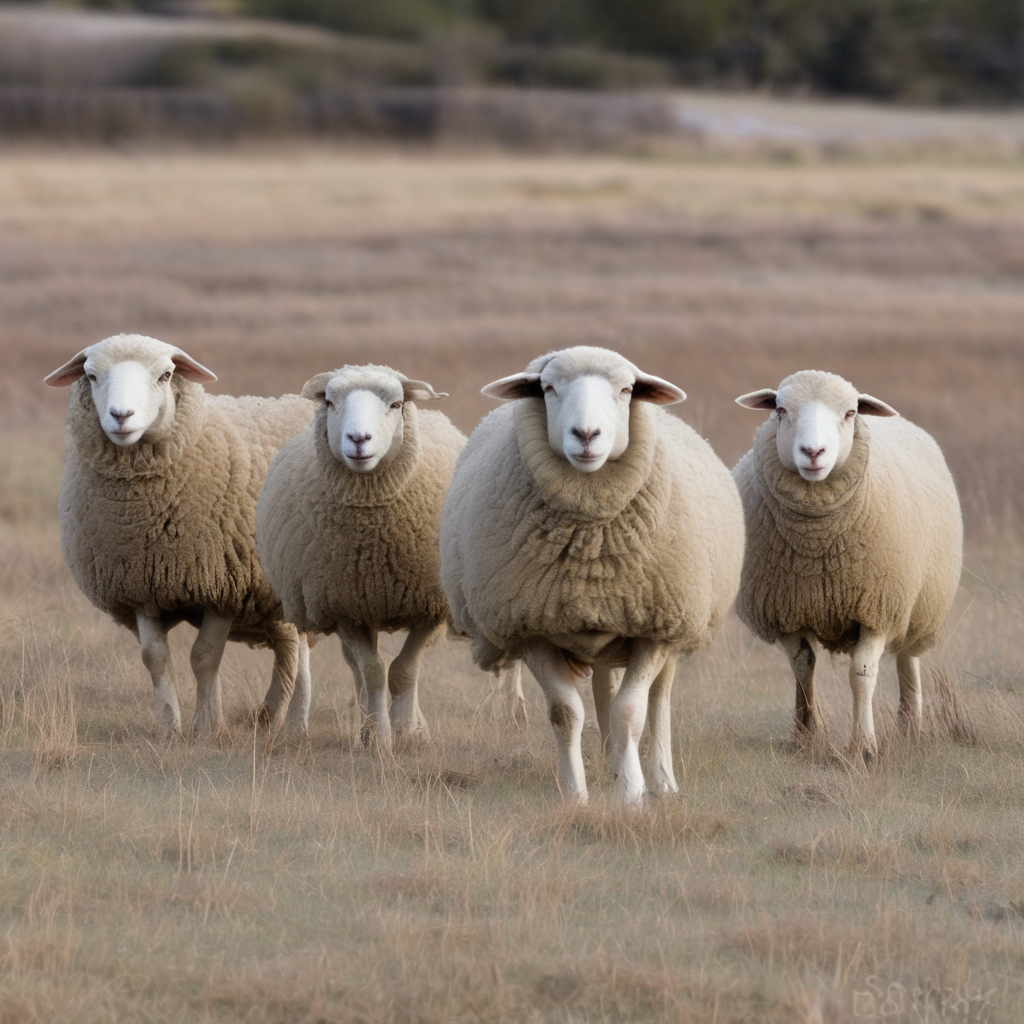}
 \end{minipage}
 
 \caption{Comparison of images generated by the original SD v1.5 \cite{ref38}, ImageReward \cite{ref15}, DDPO \cite{ref22}, and our method under the type of composition of “\textbf{Fixed Category \& Incremental quantity}”. Images in the same row are generated with the same random seed. The prompts for the sample images generated from the first row to the fourth row are: “A sheep on the prairie”, “ Two sheep on the prairie”, “Three sheep on the prairie”, and “Four sheep on the prairie”.}
 \label{fig:fc1}
\end{figure*}

\begin{figure*}[ht]
 \centering
 \begin{minipage}{0.15\textwidth}
 \centering
 \includegraphics[width=\textwidth, height=1pt]{sec/fig/1withe.png}
 \end{minipage}\hfill
 \begin{minipage}{0.187\textwidth}
 \centering
 \textbf{Stable Diffusion\cite{ref38}}
 \end{minipage}\hfill
 \begin{minipage}{0.187\textwidth}
 \centering
 \textbf{ImageReward\cite{ref15}}
 \end{minipage}\hfill
 \begin{minipage}{0.187\textwidth}
 \centering
 \textbf{DDPO\cite{ref22}}
 \end{minipage}\hfill
 \begin{minipage}{0.187\textwidth}
 \centering
 \textbf{Ours}
 \end{minipage}\\[10pt] 

 \begin{minipage}{0.15\textwidth}
 \raggedright
 Cattle on the estate.
 \end{minipage}\hfill
 \begin{minipage}{0.187\textwidth}
 \centering
 \includegraphics[width=\textwidth]{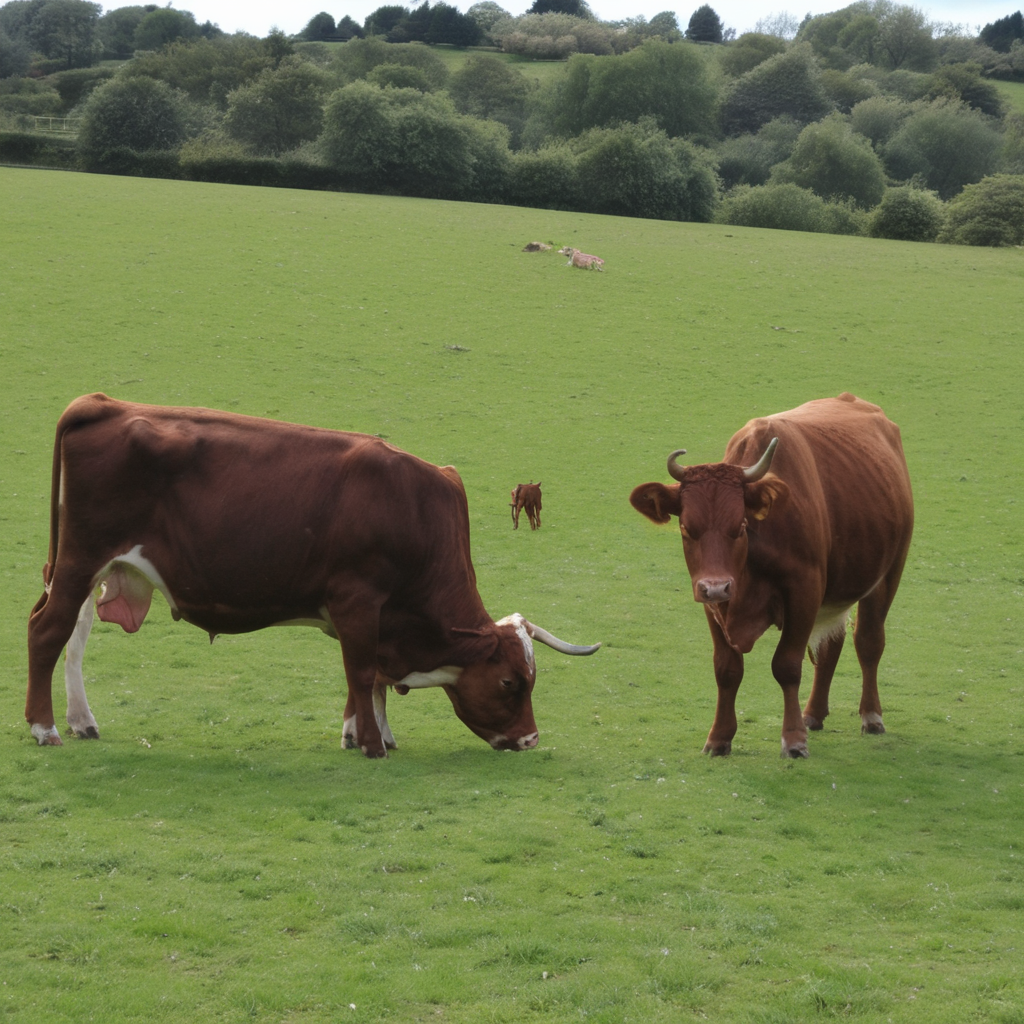}
 \end{minipage}\hfill
 \begin{minipage}{0.187\textwidth}
 \centering
 \includegraphics[width=\textwidth]{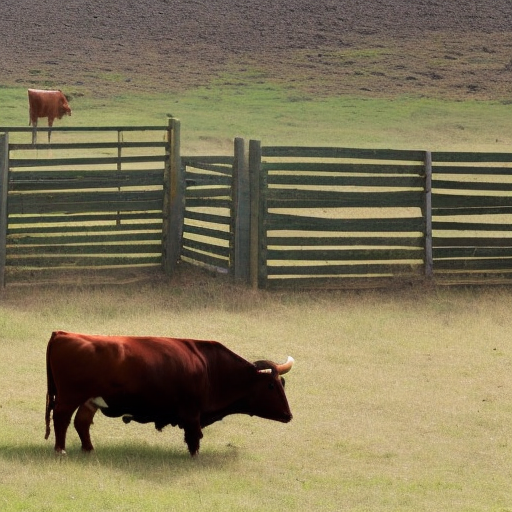}
 \end{minipage}\hfill
 \begin{minipage}{0.187\textwidth}
 \centering
 \includegraphics[width=\textwidth]{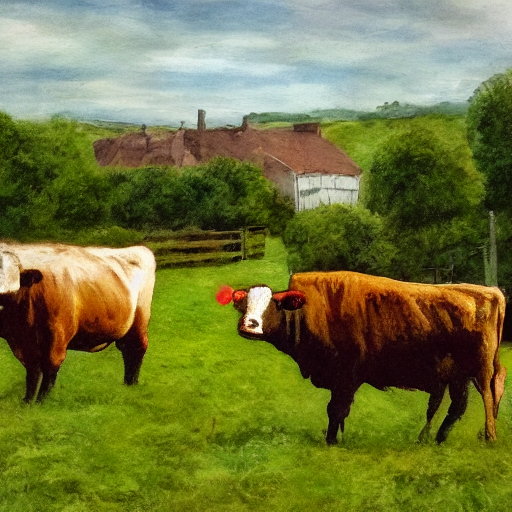}
 \end{minipage}\hfill
 \begin{minipage}{0.187\textwidth}
 \centering
 \includegraphics[width=\textwidth]{sec/fig/cp2/od1.png}
 \end{minipage}\\[10pt]

 \begin{minipage}{0.15\textwidth}
 \raggedright
 Cattle and sheep on the estate.
 \end{minipage}\hfill
 \begin{minipage}{0.187\textwidth}
 \centering
 \includegraphics[width=\textwidth]{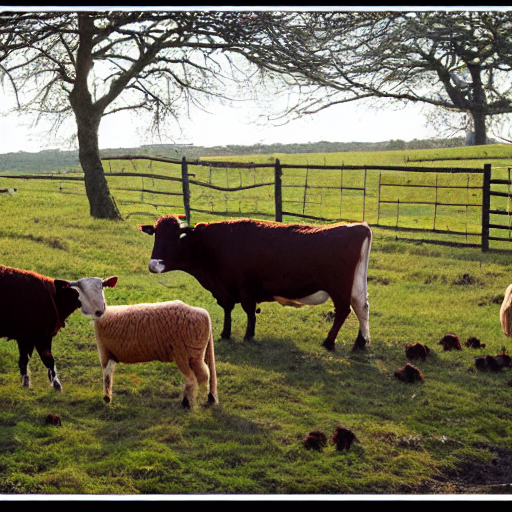}
 \end{minipage}\hfill
 \begin{minipage}{0.187\textwidth}
 \centering
 \includegraphics[width=\textwidth]{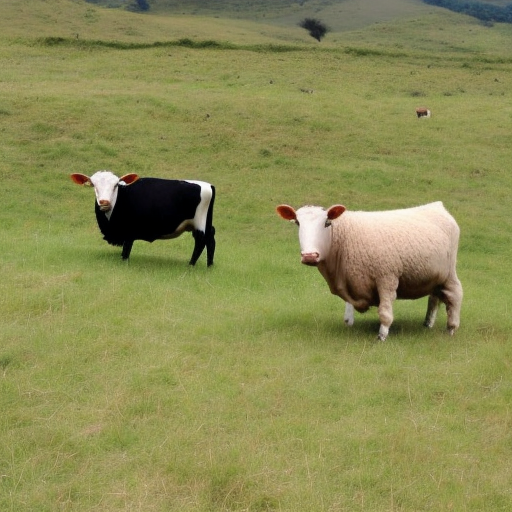}
 \end{minipage}\hfill
 \begin{minipage}{0.187\textwidth}
 \centering
 \includegraphics[width=\textwidth]{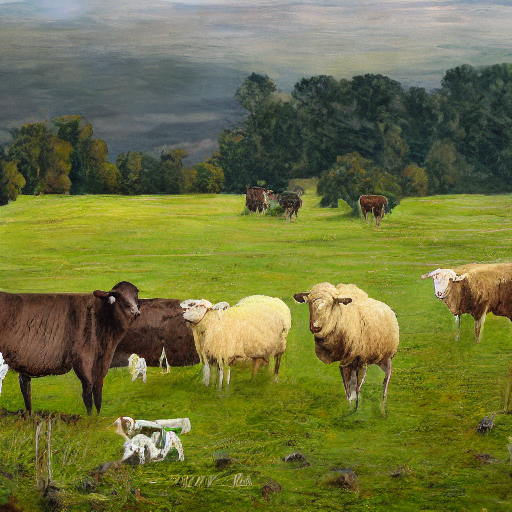}
 \end{minipage}\hfill
 \begin{minipage}{0.187\textwidth}
 \centering
 \includegraphics[width=\textwidth]{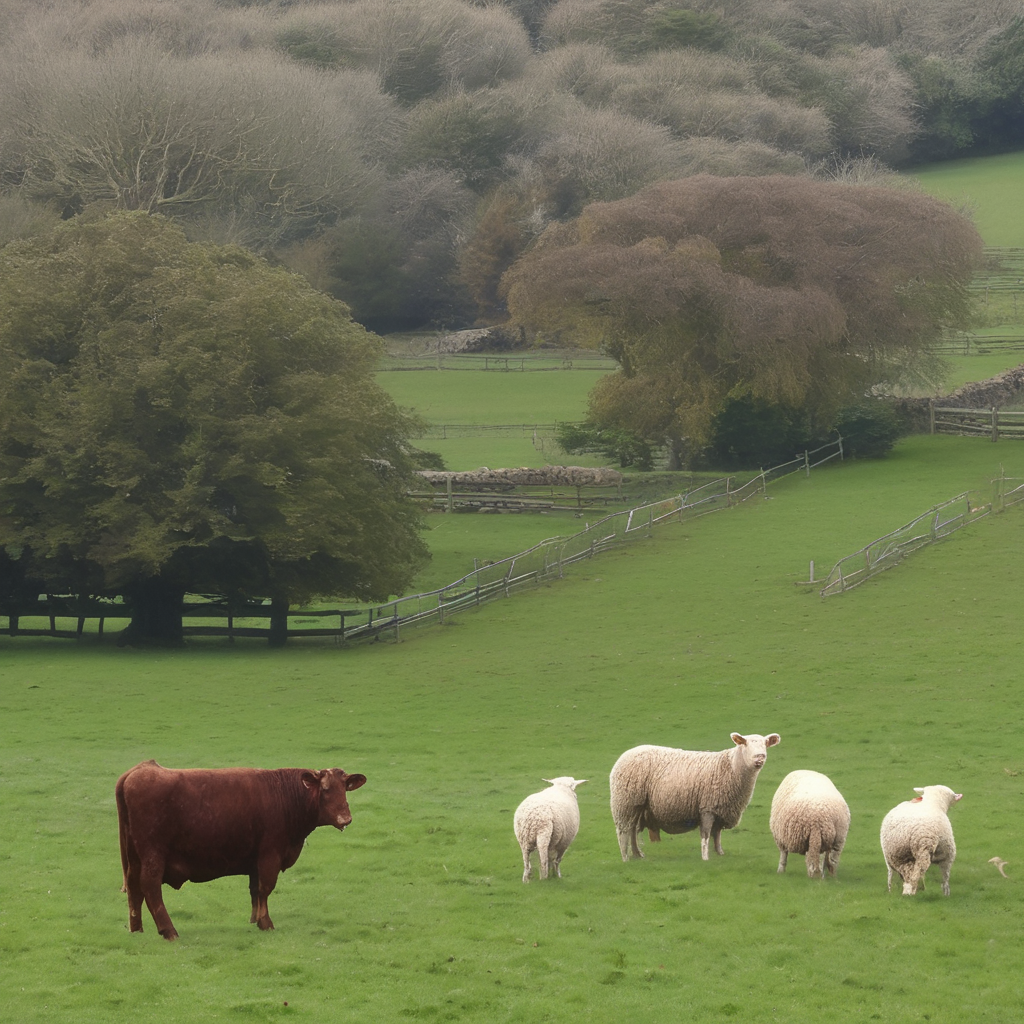}
 \end{minipage}\\[10pt]

 \begin{minipage}{0.15\textwidth}
 \raggedright
 Cattle, sheep and chicken on the estate.
 \end{minipage}\hfill
 \begin{minipage}{0.187\textwidth}
 \centering
 \includegraphics[width=\textwidth]{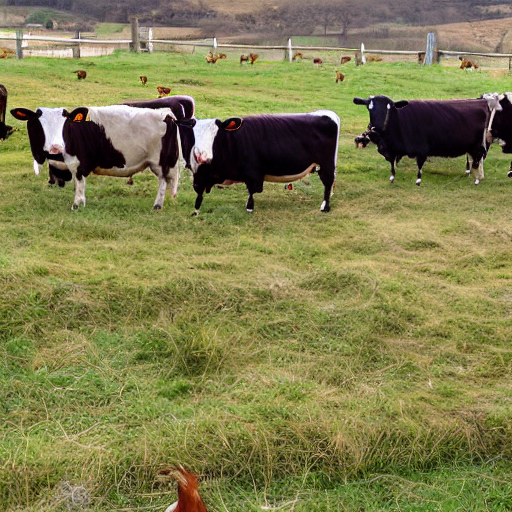}
 \end{minipage}\hfill
 \begin{minipage}{0.187\textwidth}
 \centering
 \includegraphics[width=\textwidth]{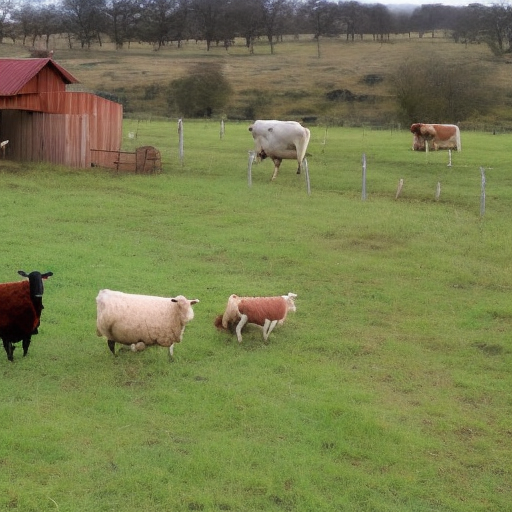}
 \end{minipage}\hfill
 \begin{minipage}{0.187\textwidth}
 \centering
 \includegraphics[width=\textwidth]{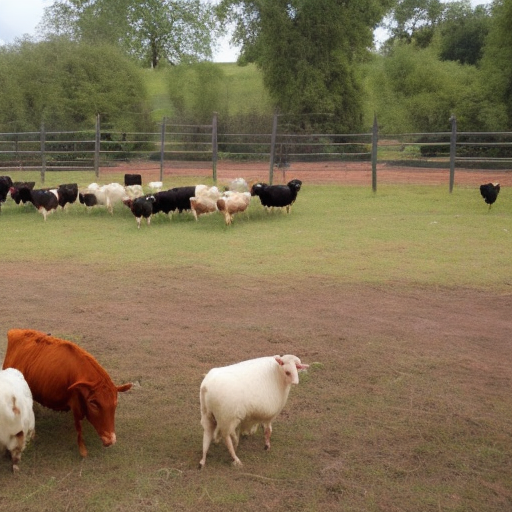}
 \end{minipage}\hfill
 \begin{minipage}{0.187\textwidth}
 \centering
 \includegraphics[width=\textwidth]{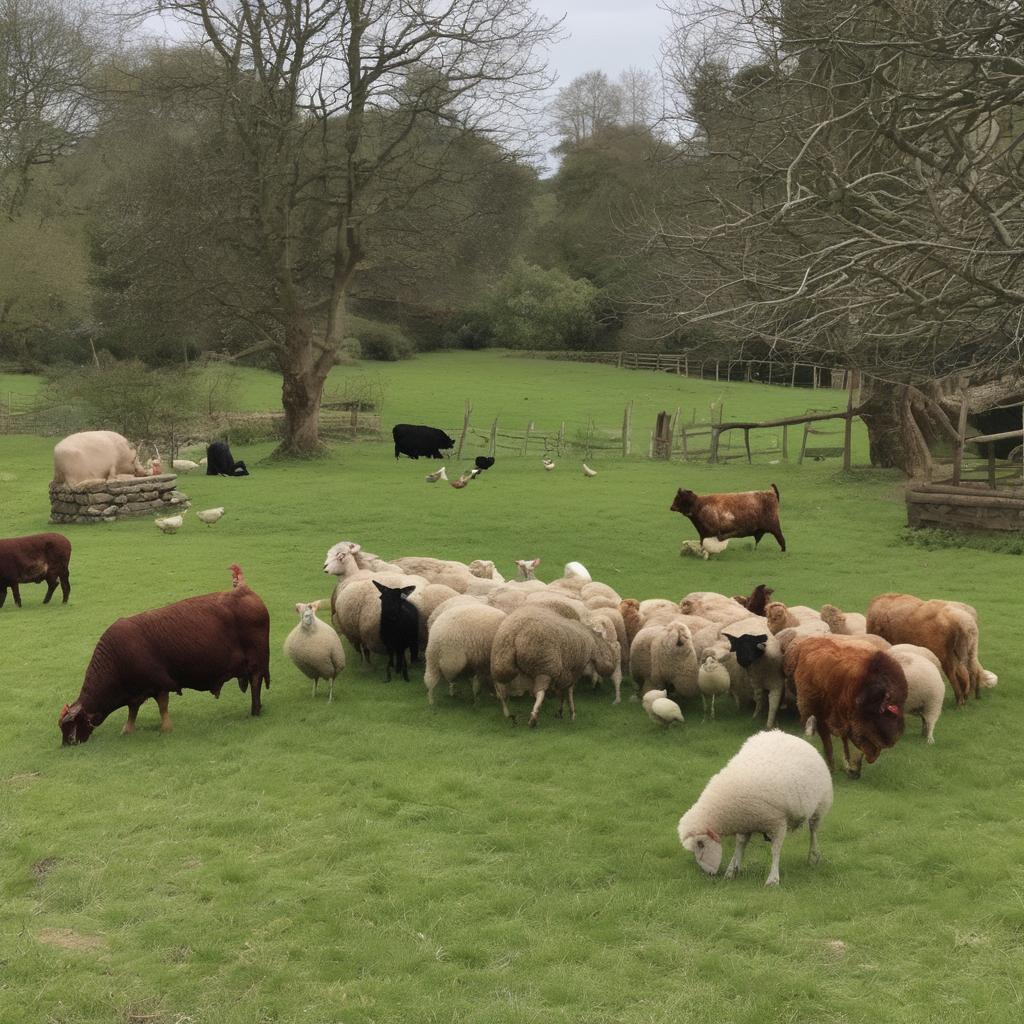}
 \end{minipage}\\[10pt]

 \begin{minipage}{0.15\textwidth}
 \raggedright
 Cattle, sheep, chicken and geese are on the estate.
 \end{minipage}\hfill
 \begin{minipage}{0.187\textwidth}
 \centering
 \includegraphics[width=\textwidth]{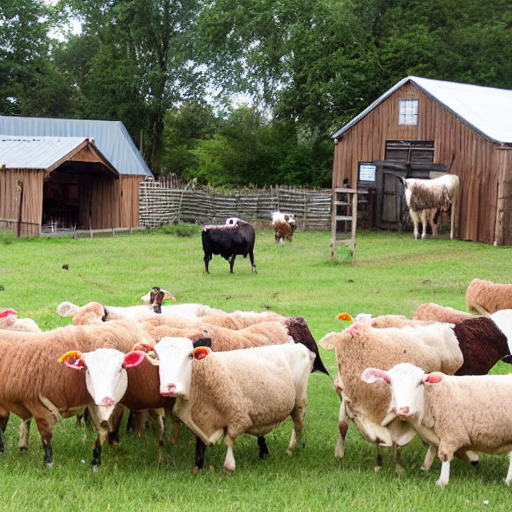}
 \end{minipage}\hfill
 \begin{minipage}{0.187\textwidth}
 \centering
 \includegraphics[width=\textwidth]{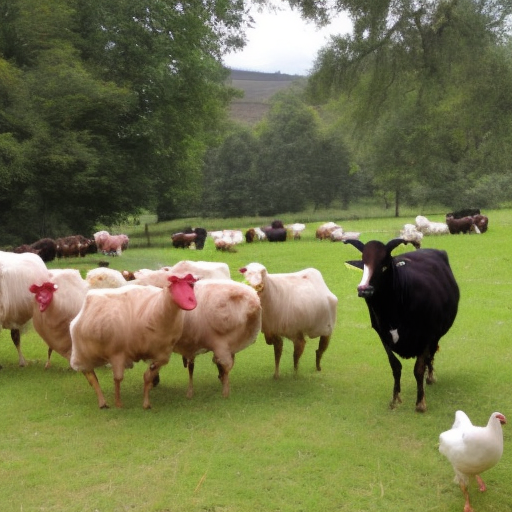}
 \end{minipage}\hfill
 \begin{minipage}{0.187\textwidth}
 \centering
 \includegraphics[width=\textwidth]{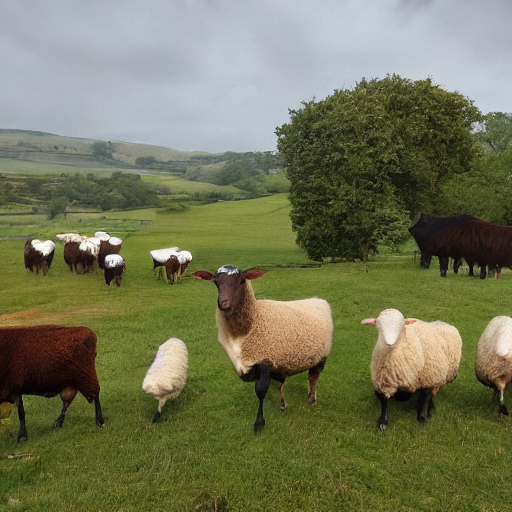}
 \end{minipage}\hfill
 \begin{minipage}{0.187\textwidth}
 \centering
 \includegraphics[width=\textwidth]{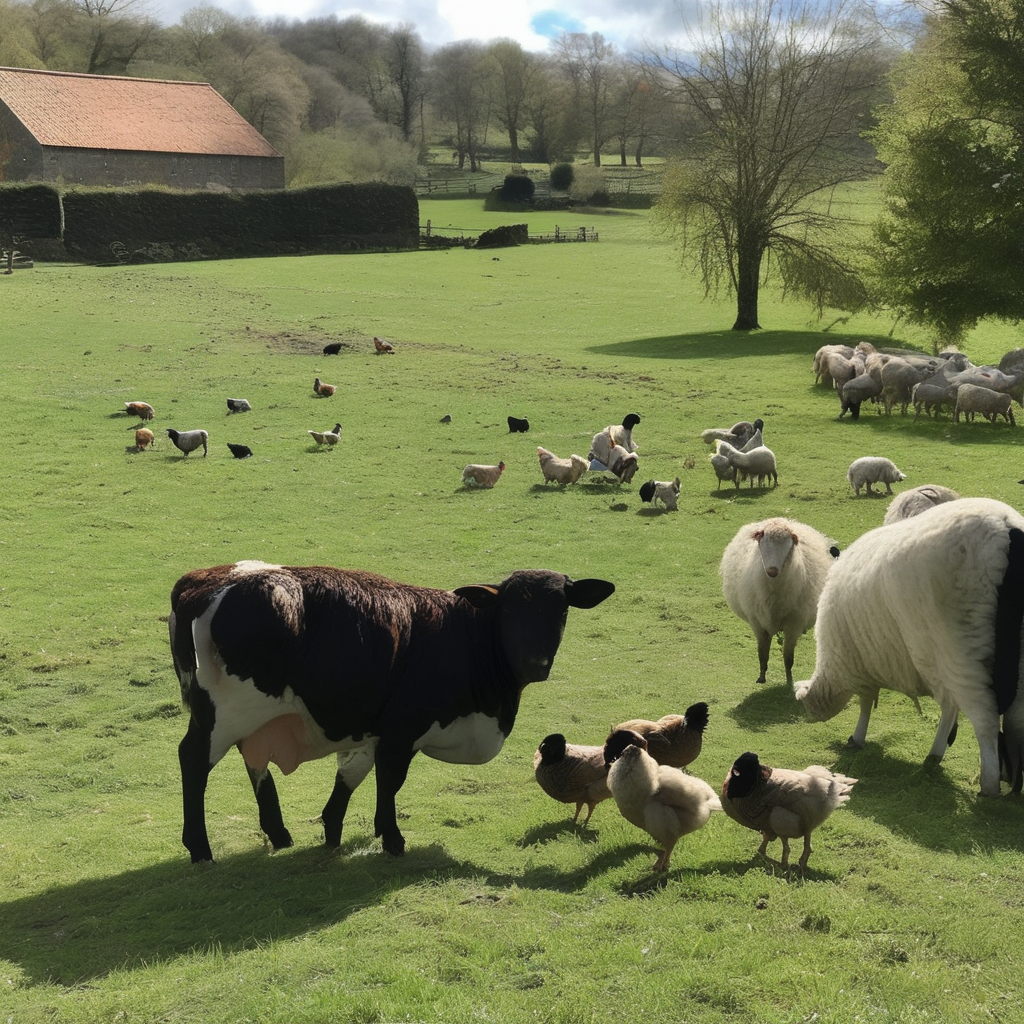}
 \end{minipage}
 
 \caption{ Comparison of images generated by the original SD v1.5 \cite{ref38}, ImageReward \cite{ref15}, DDPO \cite{ref22}, and our method under the type of composition of “\textbf{Fixed Quantity \& Incremental Category}”. Images in the same column are generated with the same random seed. The prompts for the sample images generated from the first row to the fourth row are: “Cattle on the estate”, “Cattle and sheep on the estate” , “Cattle, sheep, and chicken on the estate” , and “Cattle, sheep, chicken, and geese on the estate”.}
 \label{fig:fc2}
\end{figure*}

\begin{figure*}[ht]
 \centering
 \begin{minipage}{0.15\textwidth}
 \centering
 \includegraphics[width=\textwidth, height=1pt]{sec/fig/1withe.png}
 \end{minipage}\hfill
 \begin{minipage}{0.187\textwidth}
 \centering
 \textbf{Stable Diffusion\cite{ref38}}
 \end{minipage}\hfill
 \begin{minipage}{0.187\textwidth}
 \centering
 \textbf{ImageReward\cite{ref15}}
 \end{minipage}\hfill
 \begin{minipage}{0.187\textwidth}
 \centering
 \textbf{DDPO\cite{ref22}}
 \end{minipage}\hfill
 \begin{minipage}{0.187\textwidth}
 \centering
 \textbf{Ours}
 \end{minipage}\\[10pt] 

 \begin{minipage}{0.15\textwidth}
 \raggedright
 A horse on the prairie.
 \end{minipage}\hfill
 \begin{minipage}{0.187\textwidth}
 \centering
 \includegraphics[width=\textwidth]{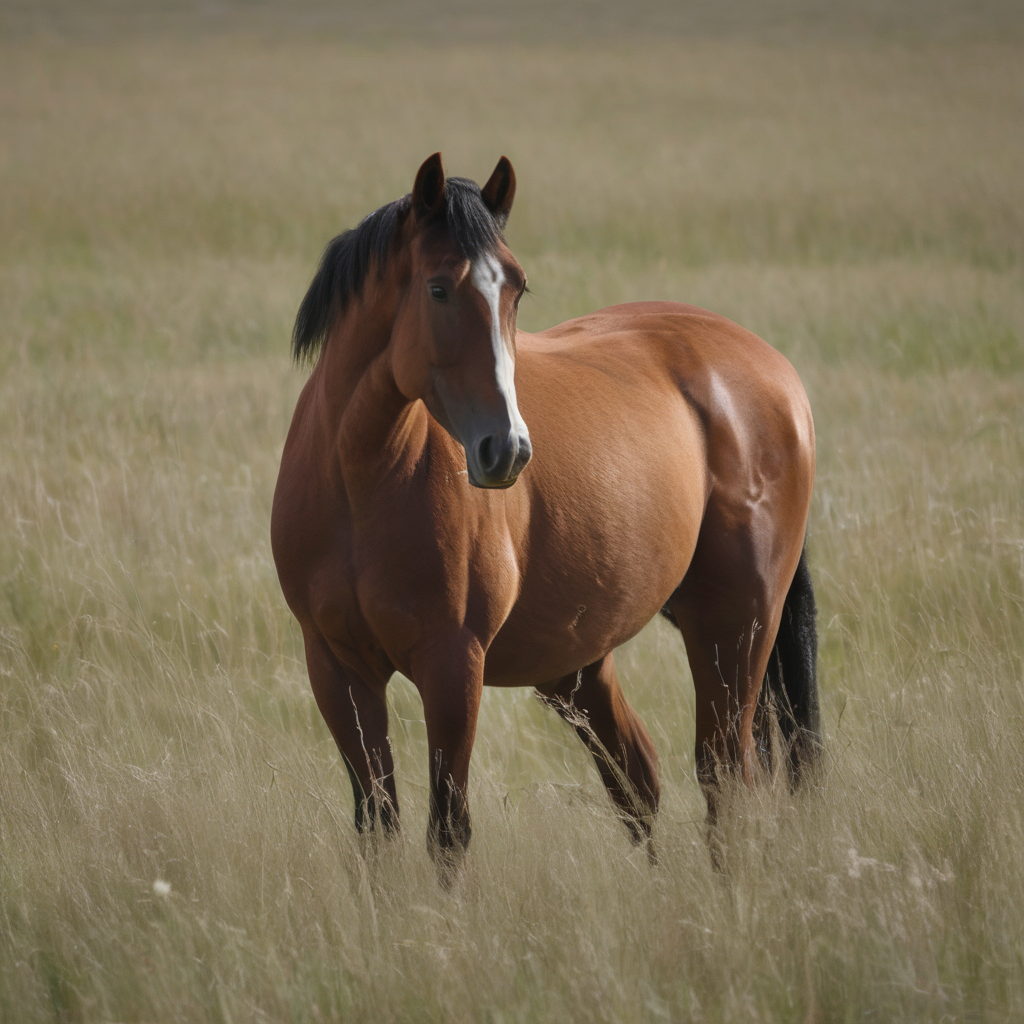}
 \end{minipage}\hfill
 \begin{minipage}{0.187\textwidth}
 \centering
 \includegraphics[width=\textwidth]{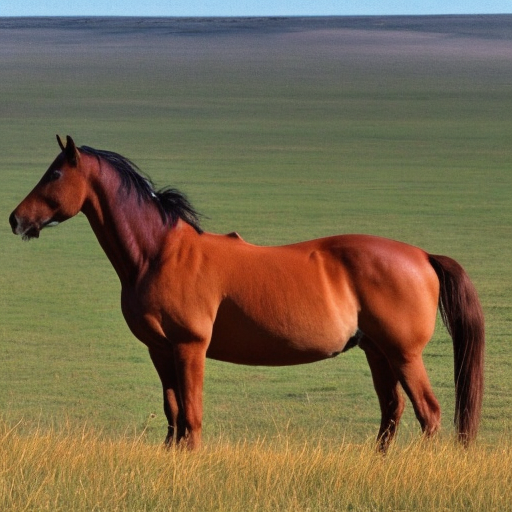}
 \end{minipage}\hfill
 \begin{minipage}{0.187\textwidth}
 \centering
 \includegraphics[width=\textwidth]{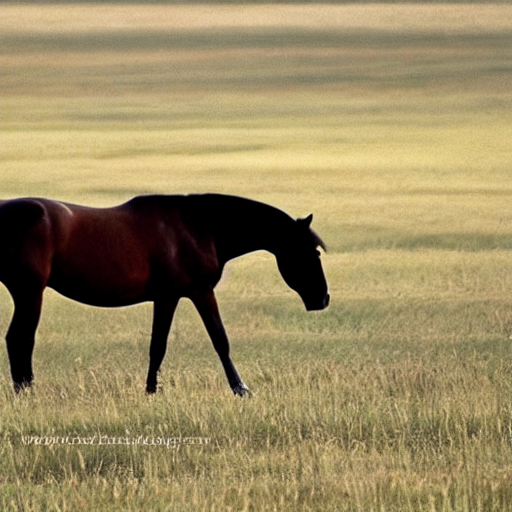}
 \end{minipage}\hfill
 \begin{minipage}{0.187\textwidth}
 \centering
 \includegraphics[width=\textwidth]{sec/fig/cp3/od1.png}
 \end{minipage}\\[10pt]

 \begin{minipage}{0.15\textwidth}
 \raggedright
 Tow horses and two sheep on the prairie.
 \end{minipage}\hfill
 \begin{minipage}{0.187\textwidth}
 \centering
 \includegraphics[width=\textwidth]{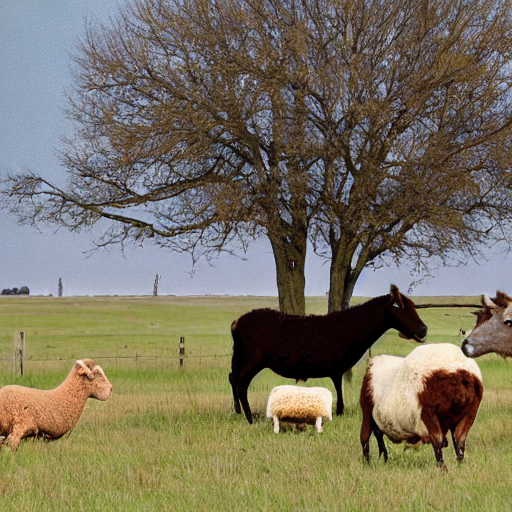}
 \end{minipage}\hfill
 \begin{minipage}{0.187\textwidth}
 \centering
 \includegraphics[width=\textwidth]{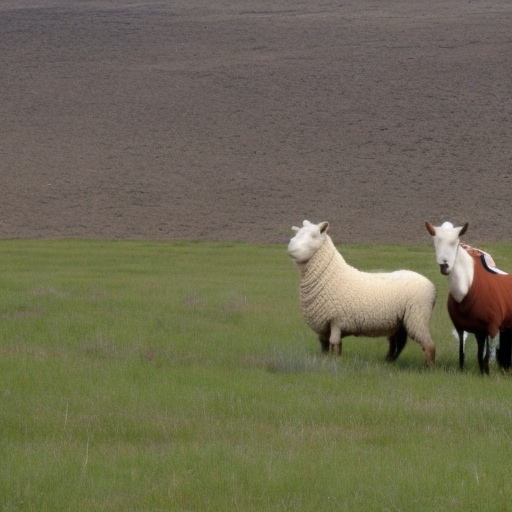}
 \end{minipage}\hfill
 \begin{minipage}{0.187\textwidth}
 \centering
 \includegraphics[width=\textwidth]{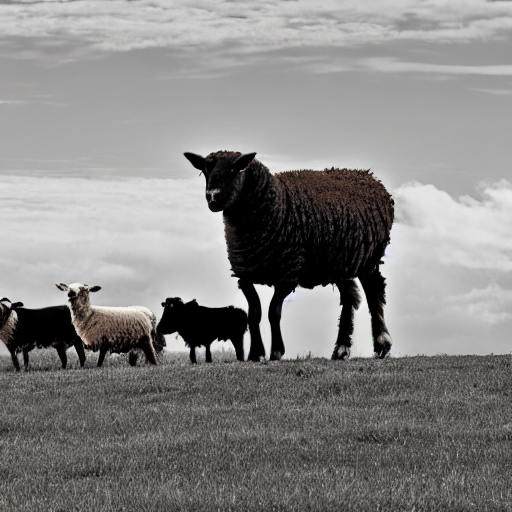}
 \end{minipage}\hfill
 \begin{minipage}{0.187\textwidth}
 \centering
 \includegraphics[width=\textwidth]{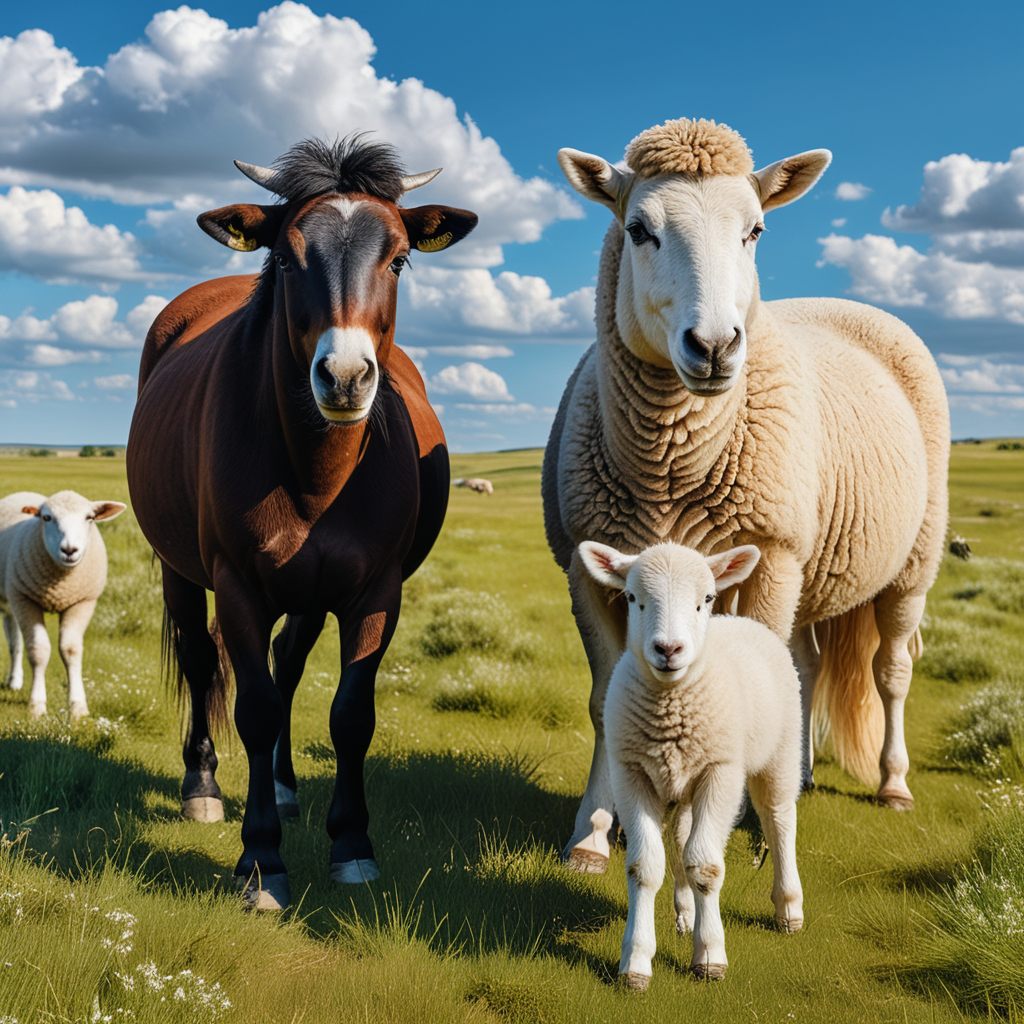}
 \end{minipage}\\[10pt]

 \begin{minipage}{0.15\textwidth}
 \raggedright
 Three horses, three cattle and three sheep on the prairie.
 \end{minipage}\hfill
 \begin{minipage}{0.187\textwidth}
 \centering
 \includegraphics[width=\textwidth]{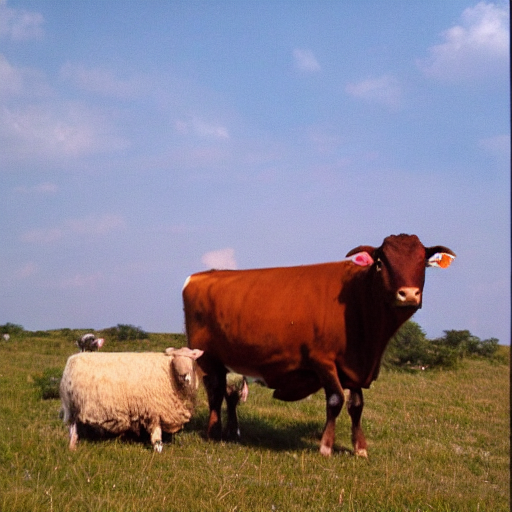}
 \end{minipage}\hfill
 \begin{minipage}{0.187\textwidth}
 \centering
 \includegraphics[width=\textwidth]{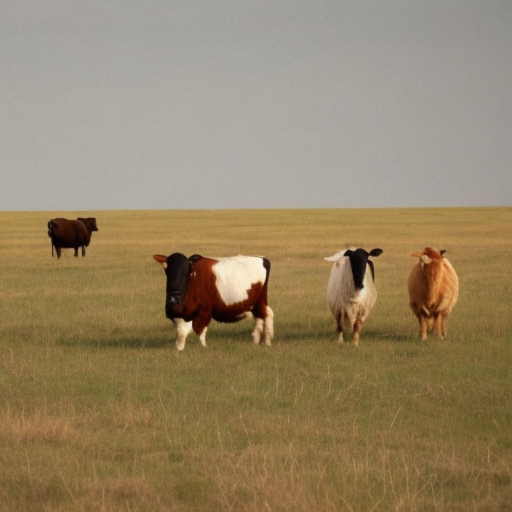}
 \end{minipage}\hfill
 \begin{minipage}{0.187\textwidth}
 \centering
 \includegraphics[width=\textwidth]{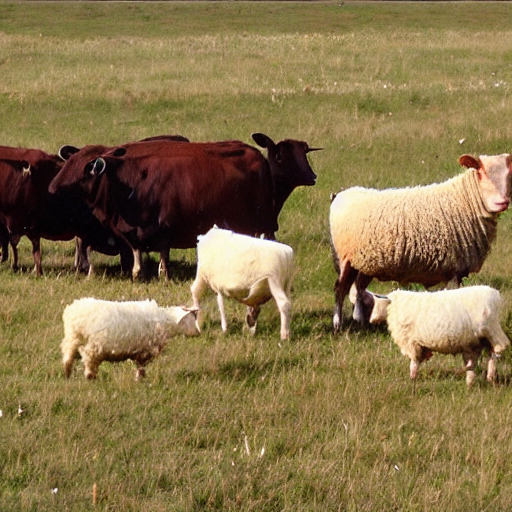}
 \end{minipage}\hfill
 \begin{minipage}{0.187\textwidth}
 \centering
 \includegraphics[width=\textwidth]{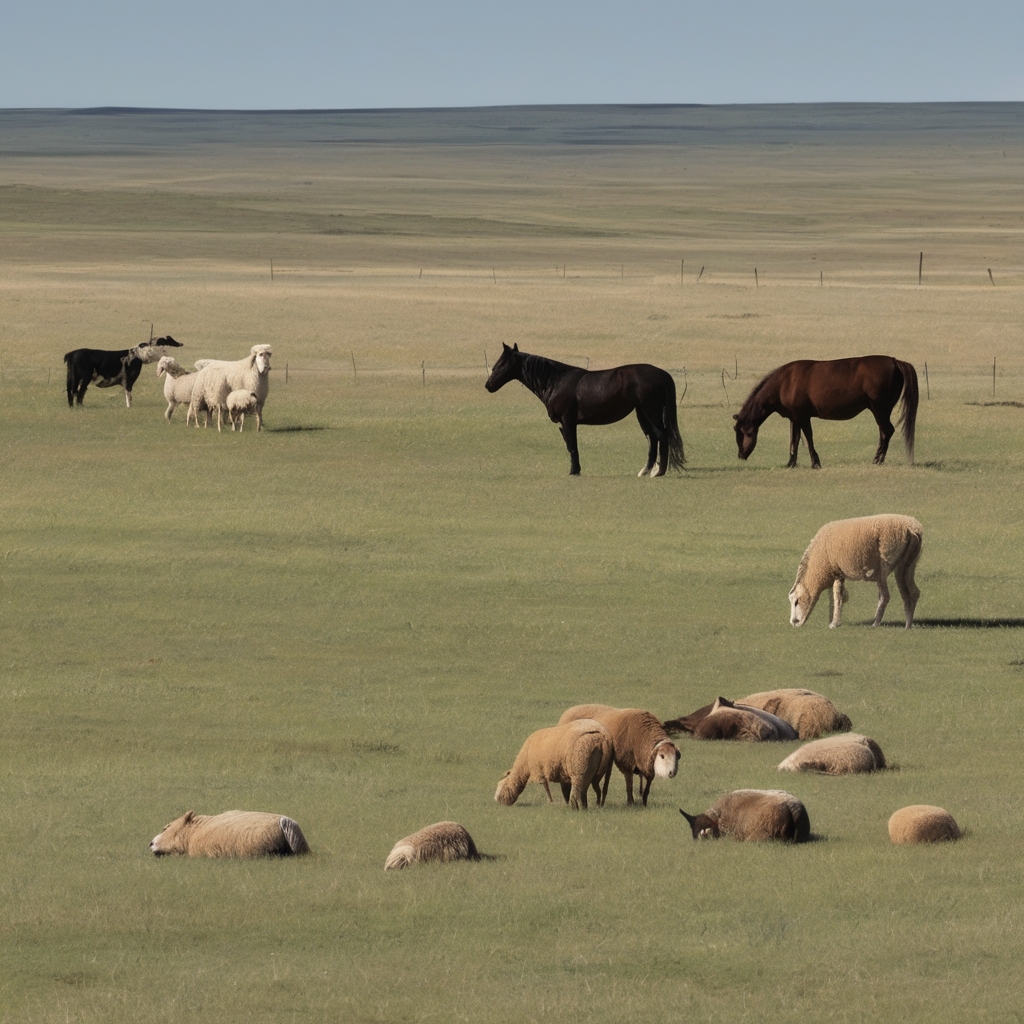}
 \end{minipage}\\[10pt]

 \begin{minipage}{0.15\textwidth}
 \raggedright
 Four horses, four cattle, four sheep and four men on the prairie.
 \end{minipage}\hfill
 \begin{minipage}{0.187\textwidth}
 \centering
 \includegraphics[width=\textwidth]{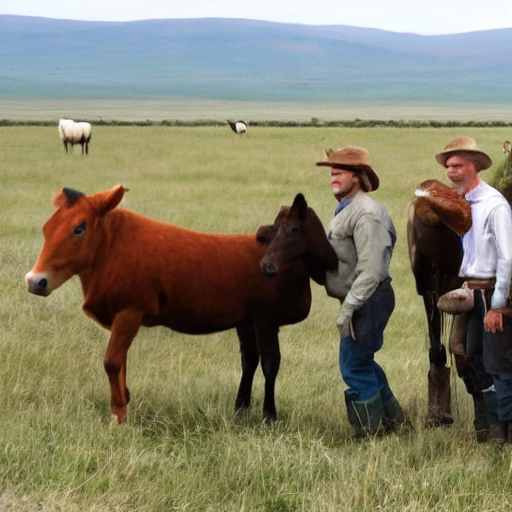}
 \end{minipage}\hfill
 \begin{minipage}{0.187\textwidth}
 \centering
 \includegraphics[width=\textwidth]{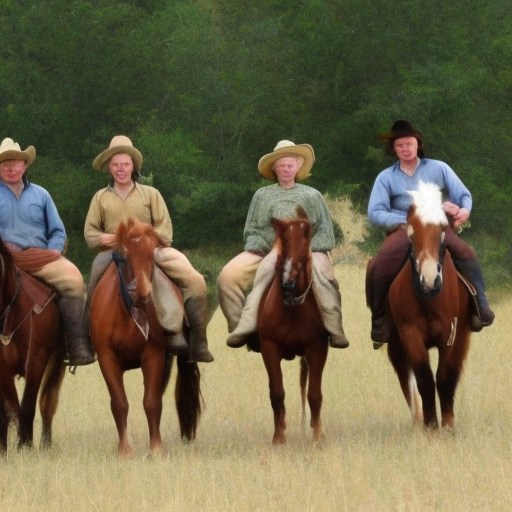}
 \end{minipage}\hfill
 \begin{minipage}{0.187\textwidth}
 \centering
 \includegraphics[width=\textwidth]{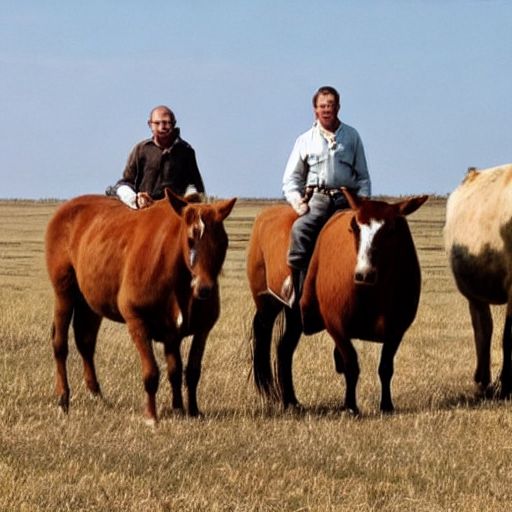}
 \end{minipage}\hfill
 \begin{minipage}{0.187\textwidth}
 \centering
 \includegraphics[width=\textwidth]{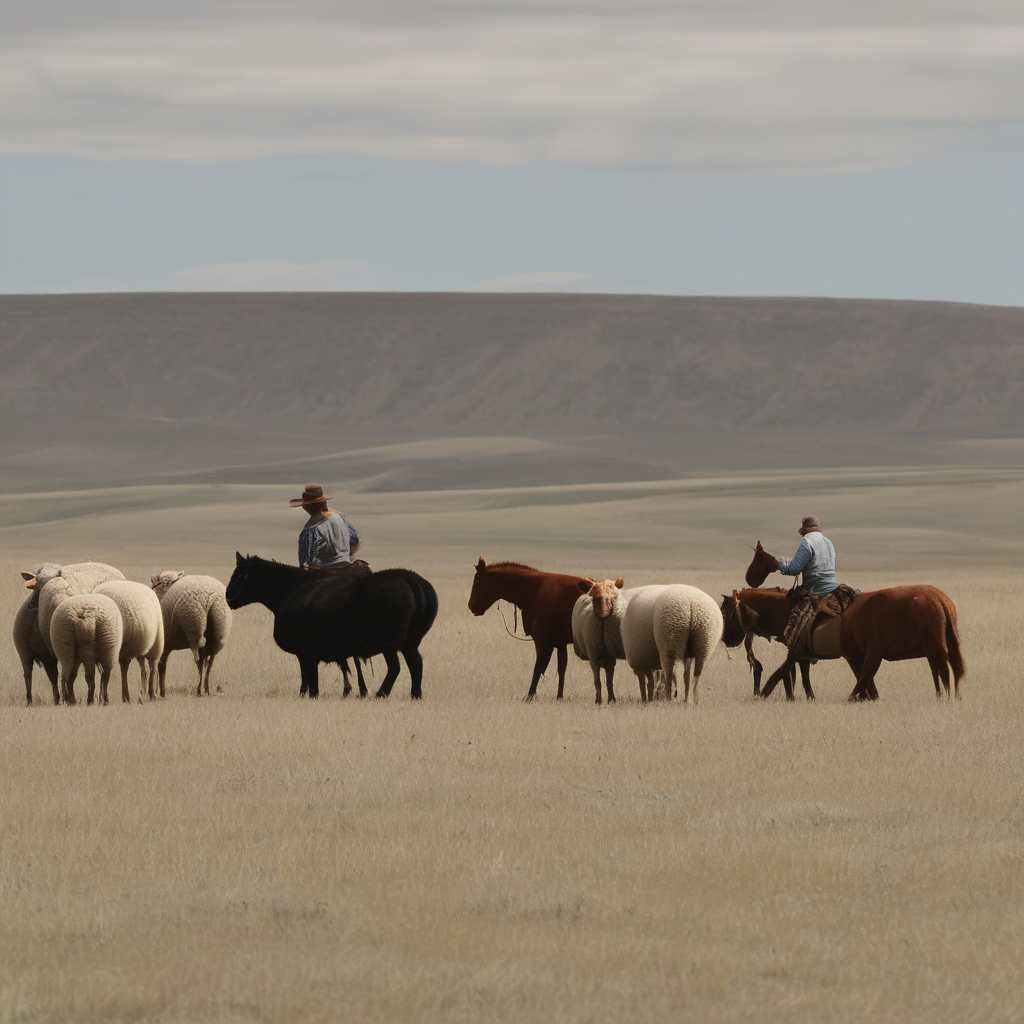}
 \end{minipage}
 
 \caption{Comparison of images generated by the original SD v1.5 \cite{ref38} , ImageReward \cite{ref15}, DDPO \cite{ref22} , and our method under the type of composition of “\textbf{Incremental Quantity \& Incremental Category}”. Images in the same row are generated with the same random seed. The prompts for the sample images generated from the first row to the fourth row are: “A horse on the prairie”, “Tow horses and two sheep on the prairie”, “Three horses, three cattle and three sheep on the prairie”, and “Four horses, four cattle , four sheep and four men on the prairie”.}
 \label{fig:fc3}
\end{figure*}

\noindent\textbf{Qualitative Comparison.} Figures \ref{fig:fc1}, \ref{fig:fc2}, and \ref{fig:fc3} present more generation results from different methods in various combinations of category and quantity. As shown in Figure \ref{fig:fc1}, \ref{fig:fc2} and \ref{fig:fc3}, our method outperforms other models in terms of both image quality and text-to-image alignment, especially in the alignment of category and quantity. 

For the type of composition of “Fixed Category \& Incremental Quantity”, although all the three comparison methods can generate images with specific category (“sheep”) and precise quantity (1, 2, 3, 4), the characteristics of the generated objects are not complete, especially when the quantity of objects is large. For example, in the image generated with DDPO \cite{ref22} model under the prompt of “Four sheep on the prairie”, one sheep lacks head (see the third column in Figure \ref{fig:fc1}).

For the type of composition of “Fixed Quantity \& Incremental Category”, all three comparison methods fail to generate the specific categories when the number of categories increase to three and four. For example, in the image generated with the ImageReward \cite{ref15} model under the prompt of “Cattle, sheep and chicken on the estate”, the category of “chicken” is missing (see the second column in Figure \ref{fig:fc2}). This can similarly be seen in the images generated with the three comparison models under the prompt of “Cattle, sheep, chicken and geese on the estate” (see the last row in Figure \ref{fig:fc2}).

For the type of composition of “Incremental Quantity \& Incremental Category”, all methods seem unable to generate images that are perfectly aligned with the prompt, but our method still generates reasonable images given complex prompt, such as “Tow horses and two sheep on the prairie” (see the last column in Figure \ref{fig:fc3}). 

Finally, while the baseline SD v1.5 \cite{ref38} often generates images with missing details (e.g., category or quantity), especially for the compositions of multiple categories and multiple quantities (see the first column of Figure \ref{fig:fc2} and Figure \ref{fig:fc3}), our model generates objects that adhere to the prompt-specified categories and quantities, as well as high quality (see the fourth column of Figure \ref{fig:fc2} and Figure \ref{fig:fc3}). We attribute the superior performance of our method to the proposed focused rewards, which emphasizes the alignment in specific category and quantity.

\subsection{Comparison results with other text-to-image models}

\noindent\textbf{Quantitative Comparison.} To further validate the advantage of the proposed model in text-image alignment, we compare our method with other types of text-to-image models, including GORS \cite{ref30} that leverages visual question answering ability of BLIP \cite{ref21} for compositional text-to-image generation, and Attend-and-Excite \cite{ref5} that leverages visual question answering ability of BLIP \cite{ref21} for evaluating attribute binding. As presented on table \ref{table:t2}, compared with the GORS \cite{ref30} and Attend-and-Excite \cite{ref5}, the CLIP Score \cite{ref19}, BLIP Score \cite{ref20} and CQ Score of our model are improved by 8.11\%, 1.34\% and 12.51\% averagely. The improvements on three metrics confirms the superiority of our model in addressing alignment problem.

\begin{table}[h!]
 \centering
 \resizebox{0.75\textwidth}{!}{ 
 \begin{tabular}{l|c|c|c}
 \hline
 Method & Clip Score ↑ & Blip Score ↑ & CQ Score ↑\\
 \hline
 Attend-and-Excite \cite{ref5} & 13.18 & 0.821 & 0.311\\
 GORS \cite{ref30} & 11.73 & 0.817 & 0.376 \\
 \hline
 \textbf{Ours} & \textbf{13.42} & \textbf{0.830} & \textbf{0.383}\\
 \hline
 \end{tabular}
 }
 \caption{Quantitative comparisons with other text-to-image models on three alignment evaluation metrics.}
 \label{table:t2}
\end{table}

\begin{figure*}[ht]
 \centering
 \begin{minipage}{0.14\textwidth}
 \centering
 \includegraphics[width=\textwidth, height=1pt]{sec/fig/1withe.png}
 \end{minipage}\hfill
 \begin{minipage}{0.247\textwidth}
 \centering
 \textbf{Attend-and-Excite\cite{ref15}}
 \end{minipage}\hfill
 \begin{minipage}{0.247\textwidth}
 \centering
 \textbf{GORS\cite{ref22}}
 \end{minipage}\hfill
 \begin{minipage}{0.247\textwidth}
 \centering
 \textbf{Ours}
 \end{minipage}\\[10pt] 


 \begin{minipage}{0.15\textwidth}
 \raggedright
 \textbf{Normal:} \\
 Four cats and two dogs resting on a sunny porch.
 \end{minipage}\hfill
 \begin{minipage}{0.247\textwidth}
 \centering
 \includegraphics[width=\textwidth]{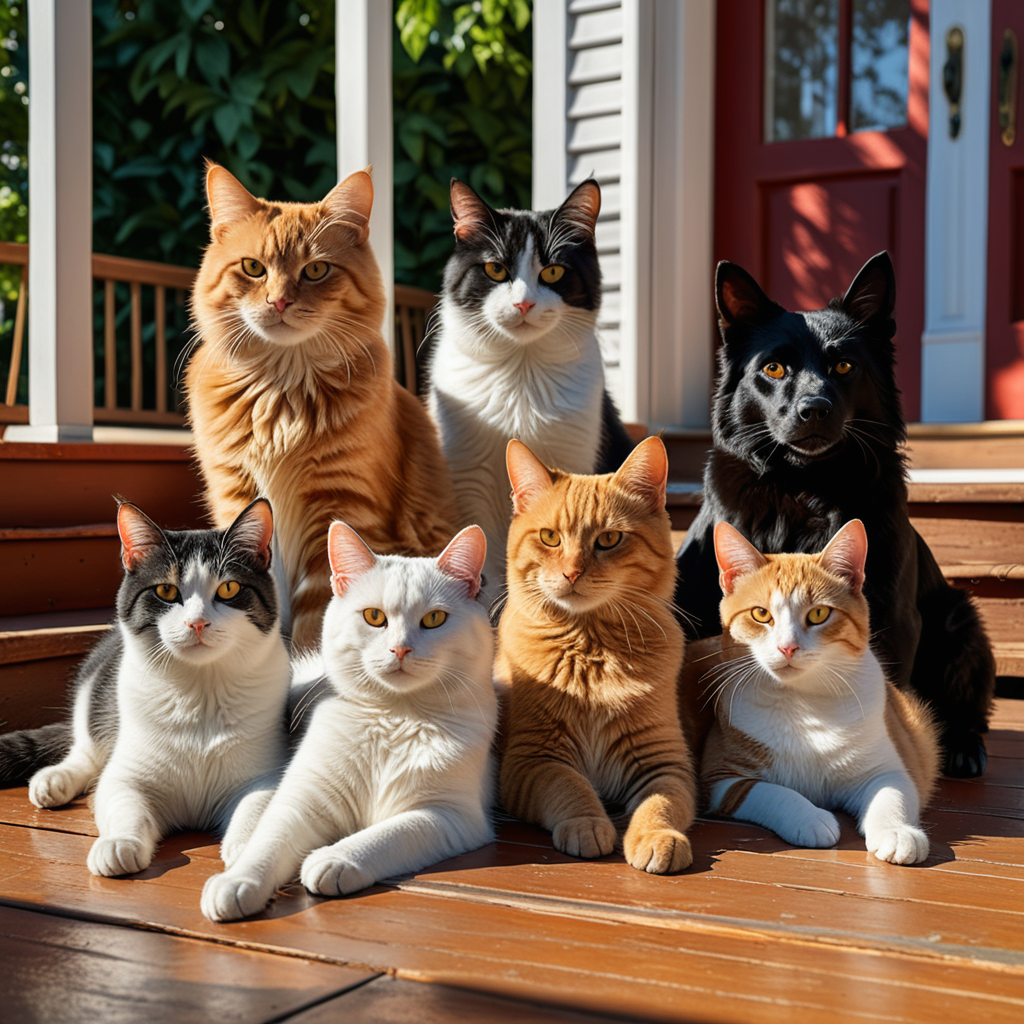}
 \end{minipage}\hfill
 \begin{minipage}{0.247\textwidth}
 \centering
 \includegraphics[width=\textwidth]{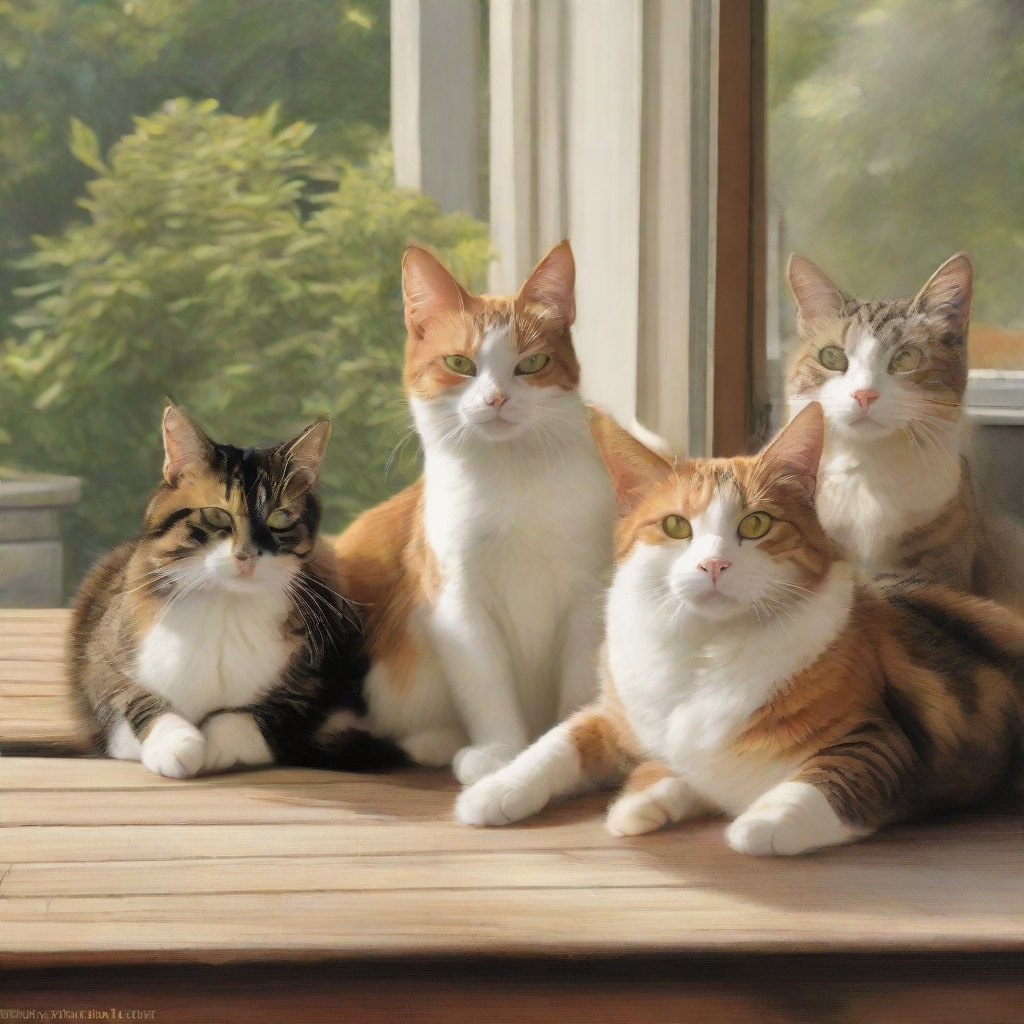}
 \end{minipage}\hfill
 \begin{minipage}{0.247\textwidth}
 \centering
 \includegraphics[width=\textwidth]{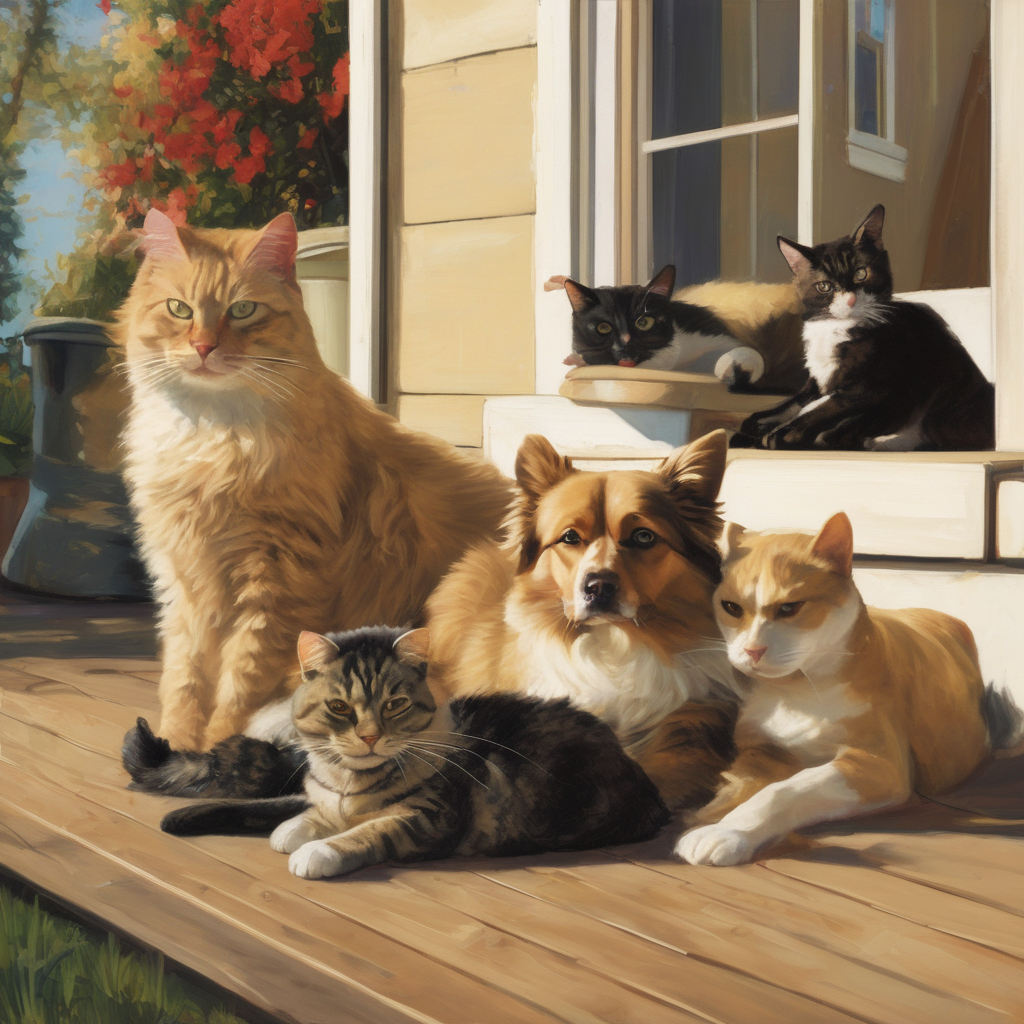}
 \end{minipage}\\[10pt]

 \begin{minipage}{0.15\textwidth}
 \raggedright
 \textbf{Awkward:} \\
 Two dogs and two cats competing in surfing at sea
 \end{minipage}\hfill
 \begin{minipage}{0.247\textwidth}
 \centering
 \includegraphics[width=\textwidth]{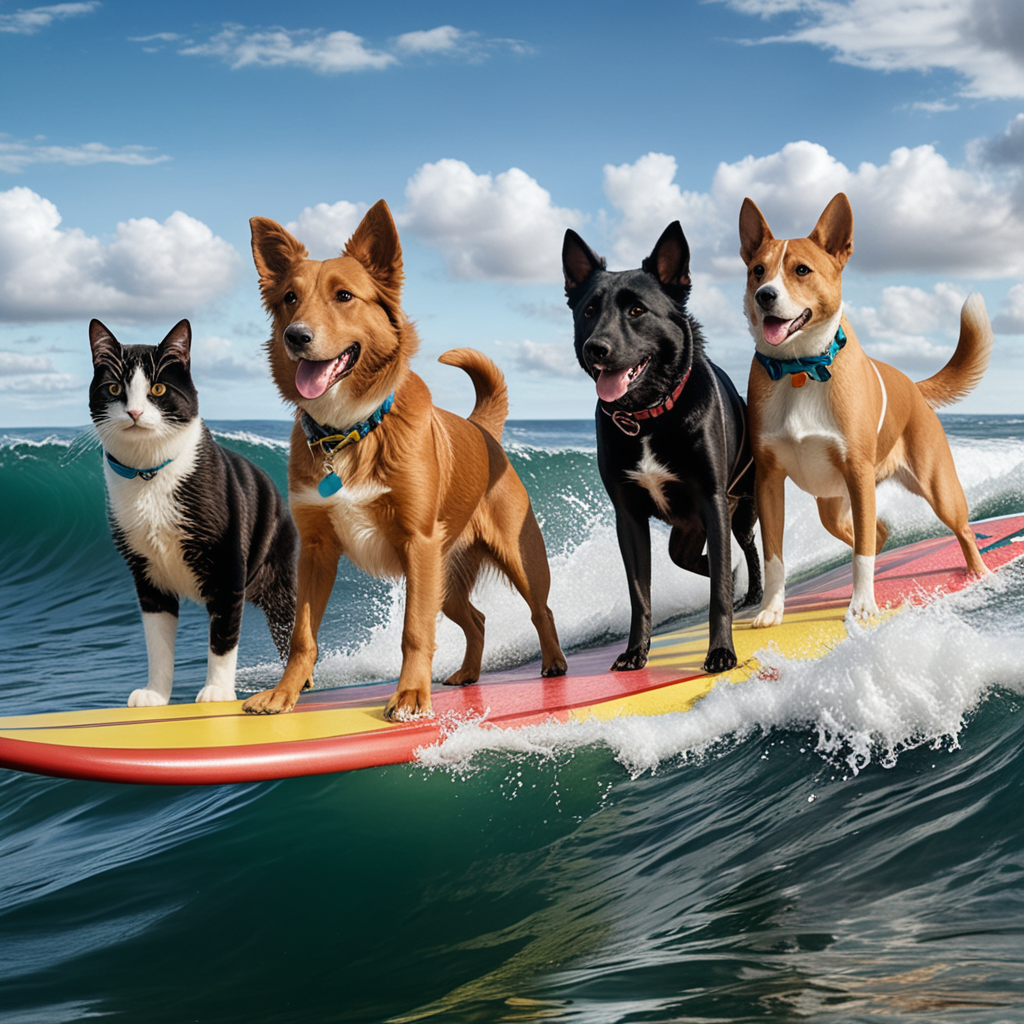}
 \end{minipage}\hfill
 \begin{minipage}{0.247\textwidth}
 \centering
 \includegraphics[width=\textwidth]{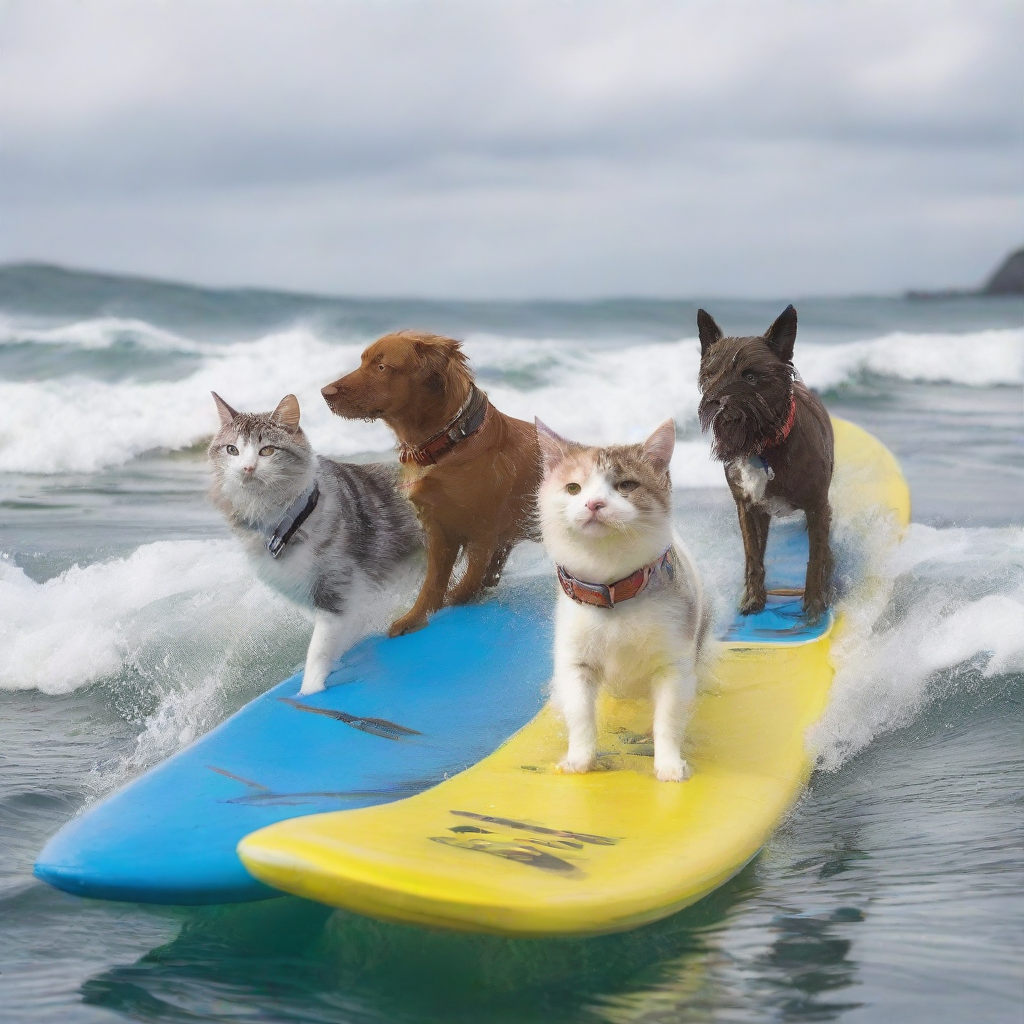}
 \end{minipage}\hfill
 \begin{minipage}{0.247\textwidth}
 \centering
 \includegraphics[width=\textwidth]{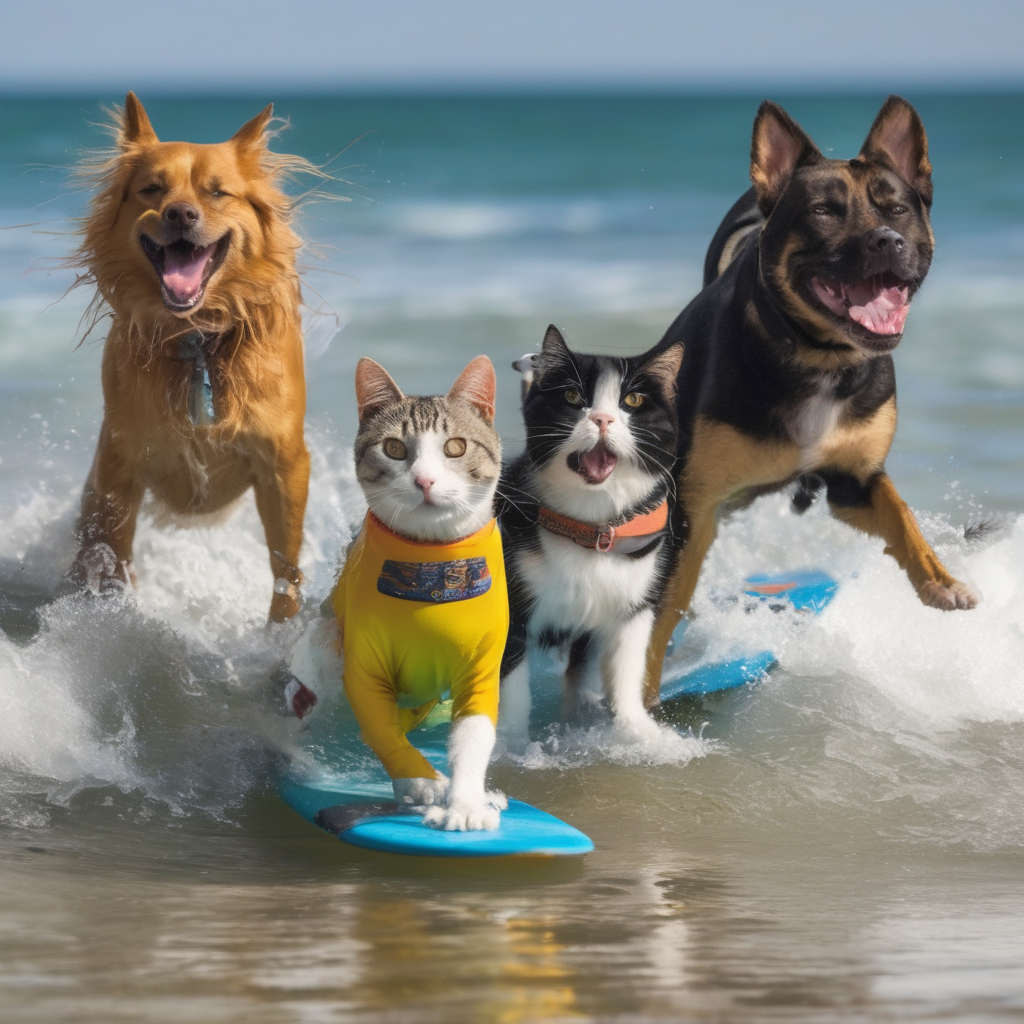}
 \end{minipage}\\[10pt]

 \begin{minipage}{0.15\textwidth}
 \raggedright
 \textbf{Unlikely:} \\
 Three polar bears walking on the moon, wearing spacesuits.
 \end{minipage}\hfill
 \begin{minipage}{0.247\textwidth}
 \centering
 \includegraphics[width=\textwidth]{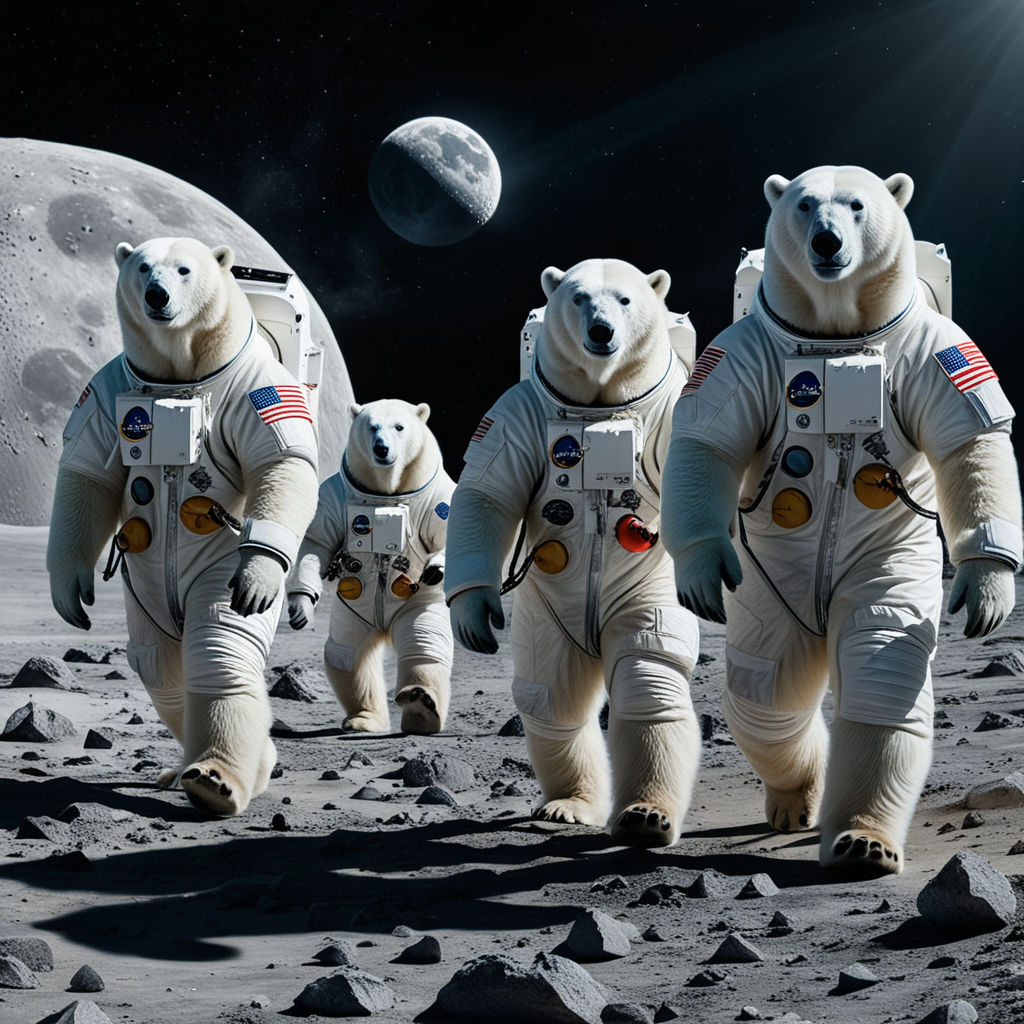}
 \end{minipage}\hfill
 \begin{minipage}{0.247\textwidth}
 \centering
 \includegraphics[width=\textwidth]{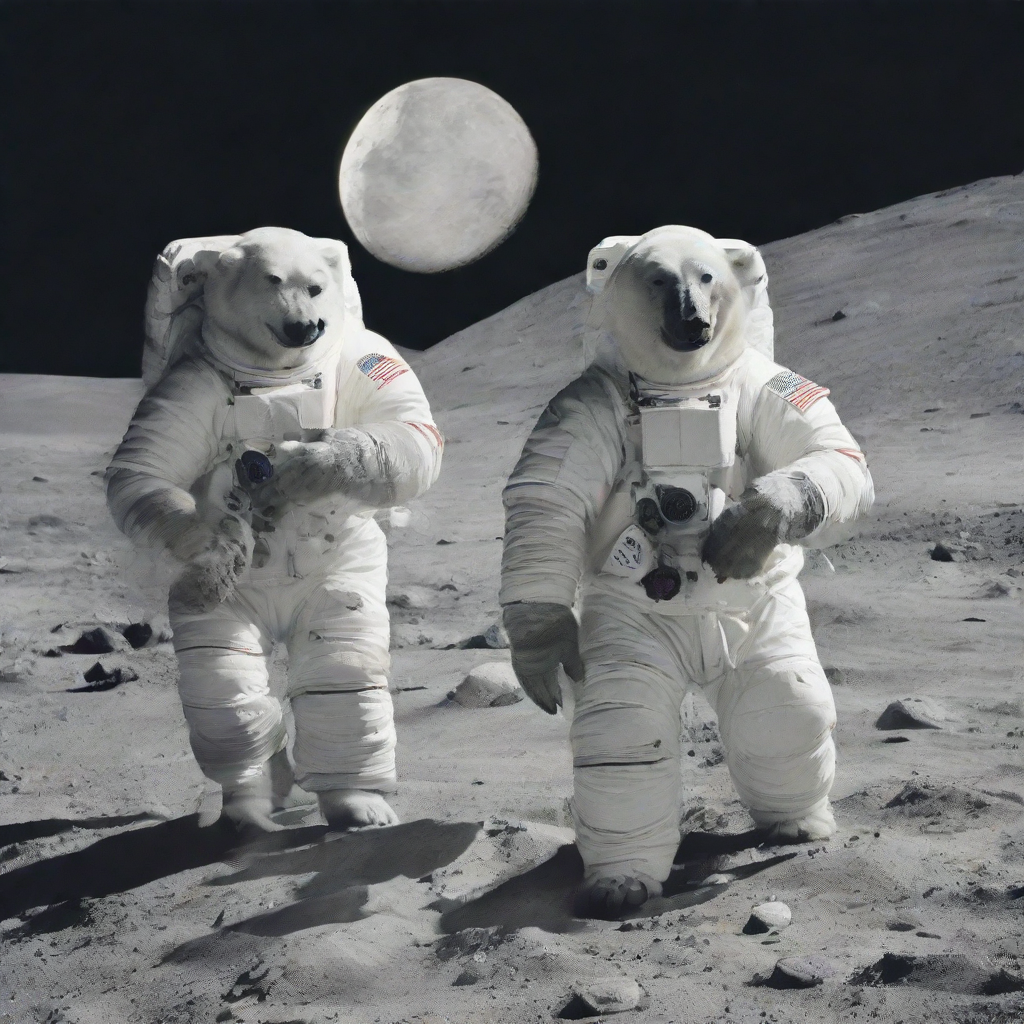}
 \end{minipage}\hfill
 \begin{minipage}{0.247\textwidth}
 \centering
 \includegraphics[width=\textwidth]{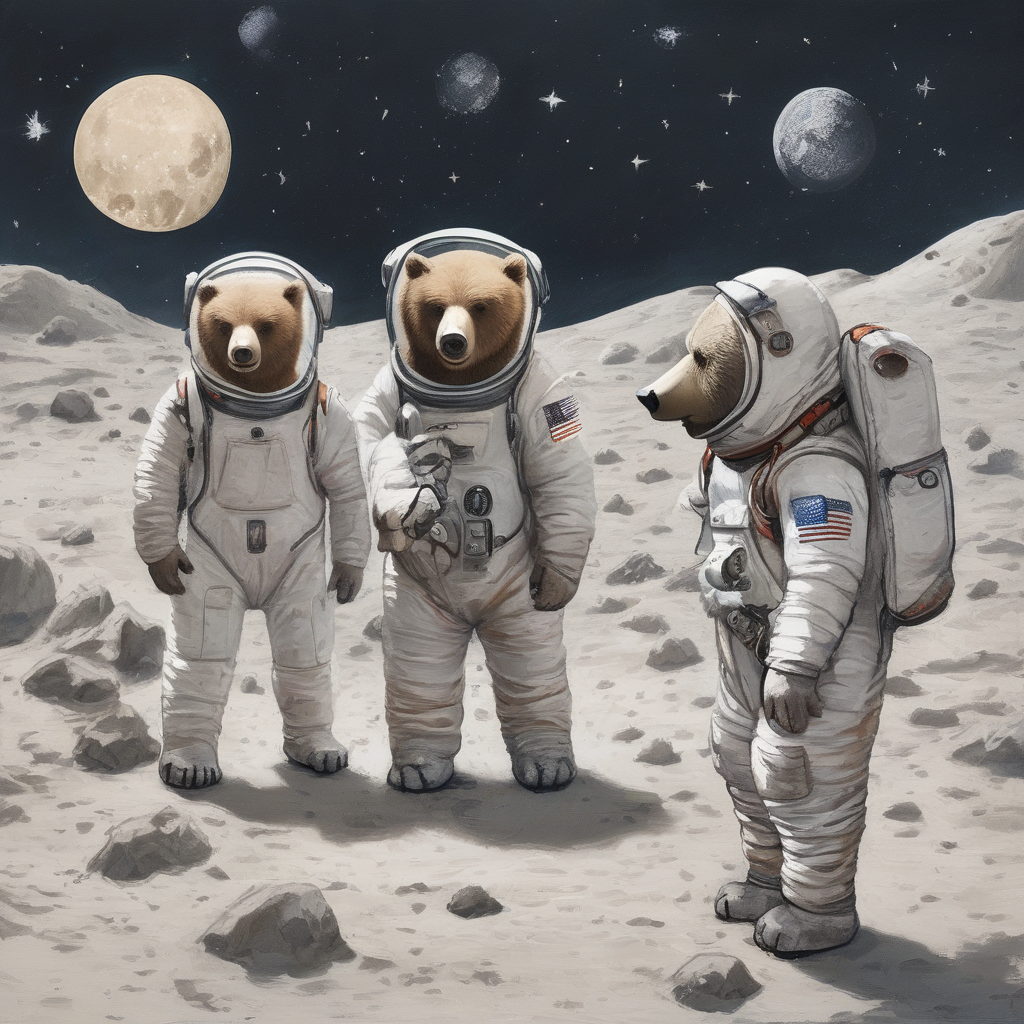}
 \end{minipage}
 
 \caption{Qualitative comparison with other methods in three kinds of compositions. \textbf{Normal} represents common composition of objects in daily life, while \textbf{Awkward} indicates the opposite. \textbf{Unlikely} refers to situations that do not exist in reality.}
 \label{fig:fc4}
\end{figure*}

\noindent\textbf{Qualitative Comparison.} Figure \ref{fig:fc4} presents more generation results against two SOTA text-to-image models: GORS \cite{ref30} and Attend-and-Excite \cite{ref5}, in three kinds of compositions mentioned in Section \ref{zw42} . As shown in Figure \ref{fig:fc4}, our method achieves a better performance in terms of both image quality and text-image alignment.

\clearpage

\renewcommand{\thesection}{B}
\section{Algorithm \label{ap2}}

\begin{algorithm*}[!h]
	\caption{Fine-tune Stable Diffusion with Feedback from Object Detector}
	\label{alg:Fine-tune}
	\renewcommand{\algorithmicrequire}{\textbf{Input:}}
	\renewcommand{\algorithmicensure}{\textbf{Output:}}
	\begin{algorithmic}[1]
		\REQUIRE Text-image pairs dataset $\mathcal{D} = \{(\bm{x}_1,\bm{z}_1),\cdots,(\bm{x}_n,\bm{z}_n)\}$; SD with pre-trained parameters $\omega_0$; SD pre-trained loss function $\psi$; reward to loss function $\varphi$; reward re-weight scale $\lambda$; the number of noise scheduler time steps $T$; the number of epochs $N$; off-the-shelf object detection model $\textit{YOLOS}$; word segmentation model $\textit{Token}$; average category confidence function $Acc$; average quantity confidence function $Aqc$; matching score function $CQScore$; learning rate $\textit{lr}_1$, $\textit{lr}_2$  
		\ENSURE Updated SD model parameter $\omega$.    
		
		\FOR{each epoch $j$ in $\textit{range}(N)$}
		\FOR{each $(\bm{x}_i,\bm{z}_i)\in{\mathcal{D}}$}
		\STATE $\mathcal{L}_{pretrain} \gets \psi(\bm{x}_i,\bm{z}_i;\omega_j^i)$ // Compute SD loss for each text-image pair $(\bm{x}_i,\bm{z}_i)$ under the current model parameter $\omega_j^i$. If $i = 1$, set $\omega_0^1 = \omega_0$
		
		\STATE $\omega_j^i \gets \omega_j^i - \textit{lr}_1 \nabla_w \mathcal{L}_{pretrain}$ // Update SD model parameter using pre-training loss
		
		\STATE $\bm{y}_T \sim \mathcal{N}(0,I)$ // Sample a noise as latent
		\FOR{$k = T,\cdots,2$}
		     \STATE \textbf{no grad:} $\bm{y}_{k-1} \gets \textit{SD}(\bm{x}_i, \bm{y}_{k};\omega_j^i)$  // Sample in latent space bm{y} reverse diffusion until $\bm{y}_1$
		\ENDFOR
		\STATE \textbf{with grad:} $\bm{y}_0 \gets \textit{SD}(\bm{x}_i,\bm{y}_1;\omega_j^i)$ // Optimize the reverse diffusion process for generating the original laten $\bm{y}_0$ from $\bm{y}_1$
		
		\STATE $\tilde{\bm{z}}_i \gets \textit{Decoder}(\bm{y}_0)$ // Transform latent to image via a Decoder
		
		\STATE $\{(\tilde{\bm{z}}_c^l: \tilde{\bm{z}} n_b^l)\}_{l = 1}^{\tilde{\bm{z}} n_c}, \{(\tilde{\bm{z}}_c^l: p_c^l)\}_{l = 1}^{\tilde{\bm{z}} n_c}\gets \textit{YOLOS}(\tilde{\bm{z}}_i)$ // Compute the quantity $\tilde{\bm{z}}n_b^l$ and confidence $p_c^l$ for each class $\tilde{\bm{z}}_c^l$ from the generated image $\tilde{\bm{z}}_i$ via \textit{YOLOS} model
		
		\STATE $\{(\bm{x}_c^l: \bm{x} n_b^l)\}_{l = 1}^{\bm{x} n_c} \gets \textit{Token}(\bm{x}_i)$ // Compute the quantity $\bm{x} n_b^l$ for each class $\bm{x}_c^l$ from text $\bm{x}_i$ via \textit{Token} modelbm{y}
		
		\STATE $Acc \gets Acc(\{(\tilde{\bm{z}}_c^l: \tilde{\bm{z}} n_b^l)\}_{l = 1}^{\tilde{\bm{z}} n_c}, \{(\tilde{\bm{z}}_c^l: p_c^l)\}_{l = 1}^{\tilde{\bm{z}} n_c},\{(\bm{x}_c^l: \bm{x} n_b^l)\}_{l = 1}^{\bm{x} n_c})$ // Compute the average category confidence for each pair of text $\bm{x}_i$ and generated image $\tilde{\bm{z}}_i$
		
		\STATE $Aqc \gets Aqc(\{(\tilde{\bm{z}}_c^l: \tilde{\bm{z}} n_b^l)\}_{l = 1}^{\tilde{\bm{z}} n_c}, \{(\bm{x}_c^l: \bm{x} n_b^l)\}_{l = 1}^{\bm{x} n_c})$ // Compute the average quantity confidence for each pair of text $\bm{x}_i$ and generated image $\tilde{\bm{z}}_i$ 
		
		\STATE $\textit{CQScore} \gets CQScore(Acc,Aqc)$ // Compute the matching score
		
		\STATE $r(\bm{x}_i,\tilde{\bm{z}}_i) \gets CQScore$ // Compute reward via matching score
		
		\STATE $\mathcal{L}_{reward} \gets \lambda \varphi(r(\bm{x}_i,\tilde{\bm{z}}_i))$ // Transform reward to loss
		
		\STATE $\omega_{j}^{i+1} = \omega_j^i - \textit{lr}_2 \nabla_w \mathcal{L}_{reward}$ // Update model parameter using reward loss
		\ENDFOR
		\STATE $\omega_{j+1}^1 = \omega_j^{n}$ // Update model parameter for next epoch
		\ENDFOR
		
	\end{algorithmic}
\end{algorithm*}

\end{document}